\documentclass[10pt,twocolumn,letterpaper]{article}
\usepackage[pagenumbers]{iccv} 

\usepackage{xcolor}         
\usepackage{amsfonts}       
\usepackage{nicefrac}       
\usepackage{microtype}      
\usepackage{amsthm}
\usepackage{bm,bbm}
\usepackage[ruled,vlined]{algorithm2e}

\newtheorem{definition}{Definition}

\usepackage{threeparttable}
\usepackage{upgreek}
\usepackage{colortbl}
\usepackage{threeparttable}
\usepackage{multirow}

\definecolor{iccvblue}{rgb}{0.21,0.49,0.74}
\usepackage[pagebackref,breaklinks,colorlinks,allcolors=iccvblue]{hyperref}

\def\paperID{2980} 
\def\confName{ICCV}
\def\confYear{2025}

\title{Robust Multi-View Learning via Representation Fusion of Sample-Level Attention and Alignment of Simulated Perturbation}

\author{
Jie Xu$^{1,3}$,
Na Zhao$^{3,*}$,
Gang Niu$^{4}$,
Masashi Sugiyama$^5$,
Xiaofeng Zhu$^{1,2,*}$
\\
{\small $^1$University of Electronic Science and Technology of China, Chengdu, China; $^2$Hainan University, Haikou, China; $^3$Singapore University } \\
{\small of Technology and Design, Singapore; $^4$Southeast University, Nanjing, China; $^5$The University of Tokyo, Tokyo, Japan}
}

\begin{document}
\maketitle

\let\thefootnote\relax\footnotetext{$^{*}$Corresponding Authors.}

\begin{abstract}
Recently, multi-view learning (MVL) has garnered significant attention due to its ability to fuse discriminative information from multiple views. However, real-world multi-view datasets are often heterogeneous and imperfect, which usually causes MVL methods designed for specific combinations of views to lack application potential and limits their effectiveness. To address this issue, we propose a novel robust MVL method (namely RML) with simultaneous representation fusion and alignment. Specifically, we introduce a simple yet effective multi-view transformer fusion network where we transform heterogeneous multi-view data into homogeneous word embeddings, and then integrate multiple views by the sample-level attention mechanism to obtain a fused representation. Furthermore, we propose a simulated perturbation based multi-view contrastive learning framework that dynamically generates the noise and unusable perturbations for simulating imperfect data conditions. The simulated noisy and unusable data obtain two distinct fused representations, and we utilize contrastive learning to align them for learning discriminative and robust representations. Our RML is self-supervised and can also be applied for downstream tasks as a regularization. In experiments, we employ it in multi-view unsupervised clustering, noise-label classification, and as a plug-and-play module for cross-modal hashing retrieval. Extensive comparison experiments and ablation studies validate RML's effectiveness. Code is available at \url{https://github.com/SubmissionsIn/RML} .
\end{abstract}

\vspace{-0.25cm}
\section{Introduction}\label{sec:introduction}

In real-world applications, algorithms usually need to handle data with multiple views or modalities in different forms, such as
multi-view data from different sensors \cite{9258396,liu2018late,panimages},
image-text and video-audio pairs in multimedia~\cite{zhao2014searching,hu2019deep,radford2021learning},
and multi-omics features in biomedical data analysis~\cite{li2023scbridge,li2025metaq}.
Compared to a single view, multiple views contain richer information and utilizing them to train more comprehensive machine learning models has given rise to a continuously intriguing research topic, \ie, multi-view learning (MVL).
The key to MVL lies in leveraging the explicit correspondences among multiple views for achieving their mutual alignment and information fusion during learning representations, and thus to enhance the performance of downstream tasks like clustering~\cite{hu2019deep,10478031}, classification~\cite{sleeman2022multimodal}, and retrieval~\cite{hu2022unsupervised}.

\begin{figure}[!t]
\centering
\begin{subfigure}{0.42\linewidth}
\includegraphics[width=\linewidth]{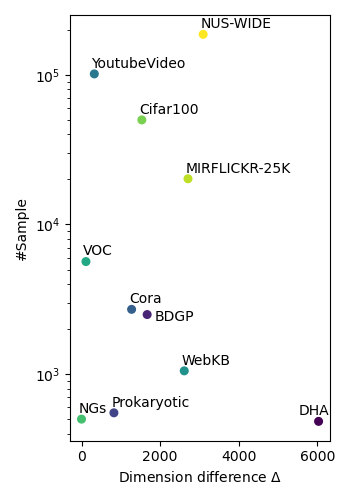}
\end{subfigure}~~
\begin{subfigure}{0.54\linewidth}
\includegraphics[width=\linewidth]{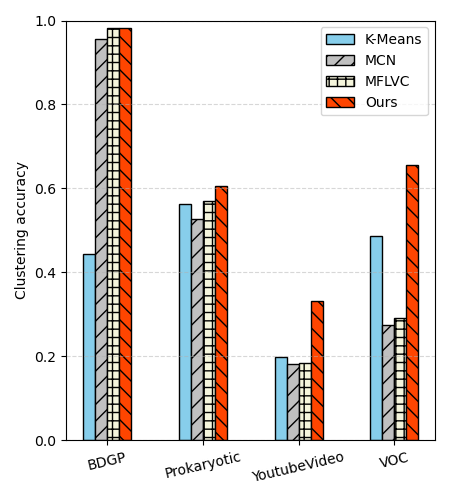}
\end{subfigure}
\vspace{-0.3cm}
\caption{Our motivation.
\emph{Left}: we utilize $\Delta = \sum_{m=1}^{V-1} |D_m-D_{m+1}|$ to measure the dimension difference across views, where $D_m$ indicates the data dimension of the $m$-th view (from Table~\ref{table0}). Real-world multi-view datasets exhibit significant differences in data modalities, dimensions, sparsity, and scales, which urges us to build view-universal and robust MVL methods.
\emph{Right}: we evaluate the performance of representation learning using an unsupervised clustering task. For example, both the methods MCN~\cite{chen2021multimodal} and MFLVC~\cite{xu2022multi} significantly outperform the baseline method K-Means~\cite{macqueen1967some} on the BDGP~\cite{cai2012joint} dataset. However, they do not make improvements on the Prokaryotic~\cite{brbic2016landscape} and YoutubeVideo~\cite{madani2013using} datasets, and even underperform K-means on the VOC~\cite{everingham2010pascal} dataset.
Our method consistently achieves good performance.}\label{moti}
\vspace{-0.3cm}
\end{figure}

\begin{figure*}[!t]
\centering
\includegraphics[width=0.99\textwidth]{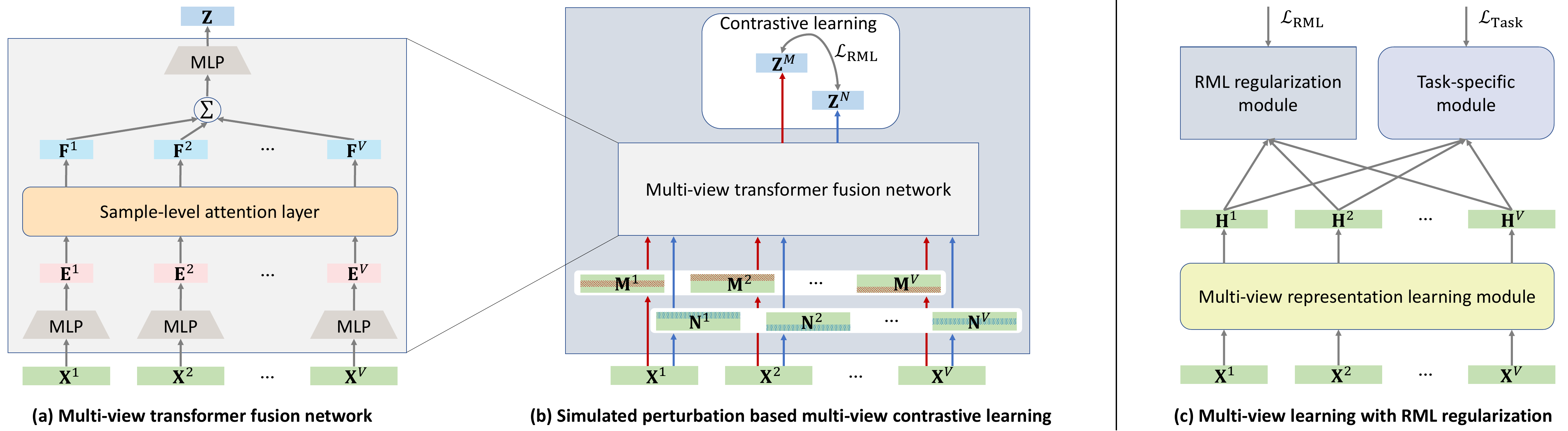}
\vspace{-0.3cm}
\caption{Our RML framework utilizes (a) and (b) for model inference and training, respectively, where (a) we propose a multi-view transformer fusion network that learns multiple homogeneous word embeddings $\{\mathbf{E}^m\}_{m=1}^V$ for multiple heterogeneous views, and then utilizes the sample-level attention across views to obtain the fused representation $\mathbf{Z}$; (b) we propose a simulated perturbation based multi-view contrastive learning which establishes the noise perturbation $\{\mathbf{N}^m\}_{m=1}^V$ and the unusable perturbation $\{\mathbf{M}^m\}_{m=1}^V$ of inputs to feed the fusion network, and the obtained $\mathbf{Z}^N$ and $\mathbf{Z}^M$ are encouraged to be aligned and discriminative by contrastive loss $\mathcal{L}_{\rm RML}$. (c) Our RML can be applied to MVL models in a plug-and-play fashion, as a regularization module to promote the specific MVL tasks.}\label{rml}
\vspace{-0.4cm}
\end{figure*}

As achieving effective information interaction across multiple views is not trivial, many existing MVL methods have been proposed by researchers and yield important progress in the past decade.
In the literature, the mainstream methodologies in MVL can be summarized into: 1) \emph{representation fusion} and 2) \emph{representation alignment}.
Specifically, to integrate multi-view discriminative information, \emph{representation fusion methods} often merge multiple views into a unified representation through multifarious fusion strategies during model construction~\cite{sleeman2022multimodal}.
For example, many MVL methods leverage the concatenation or weighted sum of multi-view representations to obtain a fused representation~\cite{zhang2024multimodal}.
Liu~\etal\cite{liu2018late} proposed a kernel-based late fusion method for clustering analysis of multi-view/modal data.
Zhang~\etal\cite{zhang2023provable} introduced a decision-level dynamic fusion method based on the multi-view energy uncertainty framework.
On the other hand, \emph{representation alignment methods} usually utilize contrastive optimization objectives to learn aligned and discriminative multi-view representations during model training~\cite{li2021align}.
For instance, a popular contrastive loss InfoNCE~\cite{oord2018representation} was widely applied in multi-view self-supervised representation learning~\cite{tian2020contrastive,radford2021learning,xu2022multi}, and these methods tend to maximize the mutual information among views for achieving their representation alignment and discrimination.
Recently, Hu~\etal\cite{10050111} further investigated the alignment problem with partially mismatched pairs in multi-view contrastive learning.

Despite the significant progress made by previous methods, the following open challenges for MVL still need to be addressed, inspiring us to explore ongoing solutions.
\emph{Firstly}, the heterogeneity of multi-view datasets challenges the universality of MVL methods.
Specifically, generalized multi-view data lack a fixed format~\cite{zhang2024multimodal}, and there are differences in data modalities, dimensions, sparsity, and scales across heterogeneous views as shown in Figure~\ref{moti}(Left). However, many methods are typically designed with specialized model structures for specific views and modalities~\cite{hu2019deep,sleeman2022multimodal}, making it difficult to apply successful experiences to other applications with different data. For example, the methods designed for multi-view~\cite{xu2022multi} or visual-audio-textual modalities~\cite{chen2021multimodal} might perform poorly on datasets comprising other views as shown in Figure~\ref{moti}(Right). The complexity of real-world applications also makes it nearly impossible to develop specialized models for arbitrary combinations of views.
\emph{Secondly}, real-world multi-view data often are imperfect that contain noise data, unusable data, and noise labels~\cite{tang2022deepicml,HanYYNXHTS18}, which demands the robustness of MVL methods.
Although some methods considered the issue of low-quality multi-view data~\cite{zhou2020end,trostenMVC,tang2022deep,zhang2023provable}, they primarily focused on balancing a small number of weights at the whole view level rather than addressing it at the finer-grained sample level.
Moreover, existing research rarely explores how to design a general MVL method which can enhance the model robustness for multiple different downstream tasks across different learning settings.
To address the aforementioned issues, we propose a novel MVL method called RML (\emph{Robust Multi-View Learning via Representation Fusion of Sample-Level Attention and Alignment of Simulated Perturbation}) as shown in Figure~\ref{rml}, which enhances the model robustness towards heterogeneous multi-view datasets and possesses the universality by the view-agnostic design to facilitate various downstream tasks.

To be specific, in model construction, we introduce a simple yet effective multi-view transformer fusion network as Figure~\ref{rml}(a).
Inspired by the fact that a sentence contains both semantic words and empty words~\cite{tang2023semantic}, we expect to correspond the usable and unusable views in a multi-view sample to the semantic and empty words in a sentence, respectively.
Therefore, our RML first establishes multilayer perceptrons (MLPs) to convert heterogeneous views into homogeneous word embeddings~\cite{vaswani2017attention}.
Then, for each multi-view sample, RML utilizes the sample-level attention layer to explore the dependencies among multiple views and output the encoded embeddings.
To capture the discriminative information among all views for the sample, RML sums all encoded embeddings to obtain a fused representation.

In model optimization, we propose a simulated perturbation based multi-view contrastive learning framework as Figure~\ref{rml}(b).
Concretely, RML generates two perturbed versions of the multi-view data by adding noise and discarding portions on random views of each sample, respectively simulating noisy and unusable data in real-world imperfect scenarios.
The two different perturbed multi-view data generate two distinct fused representations through the shared fusion network.
Subsequently, RML performs contrastive learning (with the InfoNCE loss~\cite{oord2018representation}) between them for representation alignment, to make the model robust to dynamic perturbations as well as explore the hidden discriminative information.
In this novel way, RML simultaneously achieves multi-view representation fusion and alignment.

RML can conduct self-supervised multi-view representation learning alone, and in this case Figures~\ref{rml}(a) and (b) show its model inference and training processes, respectively.
Moreover, as shown in Figure~\ref{rml}(c), RML can serve as a regularization to enhance downstream tasks when we take the hidden representations of other deep MVL methods as the input.
Our contributions are summarized as follows:
\begin{itemize}
    \item Different from previous weighting strategies at the view-level, we propose a sample-level attention based multi-view representation fusion framework that generates self-attention scores on each sample's multiple views for fine-grained fusion. This helps address the issue of the imperfect cases in real-world heterogeneous multi-view data.
    \item We introduce a simulated perturbation based contrastive learning method to train the multi-view transformer fusion network. The alignment between simulated perturbations facilitates the information interaction and representation discrimination among multiple views, and increases the model robustness to noisy, unusable data and noise labels.
    \item Unlike previous MVL methods that were usually dedicated to one specific task, our proposed RML is with universality and helps to increase the application potential of MVL. RML was employed on multi-view clustering, noise-label multi-view classification, cross-modal retrieval tasks, and extensive experiments demonstrated its effectiveness.
\end{itemize}

\section{Method}\label{MM}

In this section, we introduce the model framework, training objective, and regularization function of RML.

\subsection{Multi-view transformer fusion network}
Multi-view fusion is an abstract concept with various implementation schemes in multi-view learning, unified by the goal of extracting discriminative information from multiple views for downstream tasks.
We first provide a formal definition of multi-view fusion and then introduce our model.

\begin{definition}[Multi-View Fusion]
Given a multi-view dataset $\{\mathbf{X}^m \in \mathbb{R}^{N \times D_m}\}_{m=1}^V$ consisting of $N$ samples from $V$ views, where $D_m$ denotes the data dimension of the $m$-th view, the multi-view fusion is defined as a function $\mathcal{F}$:
\begin{equation}\label{fusion}
\mathbf{Z}= \mathcal{F}_{\theta_f}(\mathbf{X}^1,\mathbf{X}^2,\dots,\mathbf{X}^V),
\end{equation}
where $\theta_f$ is the trainable parameter of the multi-view fusion model, $\mathbf{Z} \in \mathbb{R}^{N \times d}$ is the fused representation of the dataset, and $d$ is the dimension of the fused representation.
\end{definition}

In the paradigm of Eq.~(\ref{fusion}), the most common approach to implement the multi-view fusion is utilizing deep neural networks and a view-level weighting strategy through multiple weights. This approach can be formulated as follows:
\begin{equation}\label{Wfuison}
\mathbf{Z}= \mathcal{F}_{\theta_f}(\{w^m,\mathbf{E}^m\}_{m=1}^V) = \mathcal{F}_{\theta_f}(\{w^m, \mathcal{P}_{\theta^m}(\mathbf{X}^m)\}_{m=1}^V),
\end{equation}
where $\mathbf{Z} \in \mathbb{R}^{N \times d}$, $w^m \in \mathbb{R}$, $\mathbf{E}^m \in \mathbb{R}^{N \times d_m}$, $\mathbf{X}^m \in \mathbb{R}^{N \times D_m}$.
$\mathcal{P}_{\theta^m}(\mathbf{X}^m)$ is a deep neural network that projects the input data $\mathbf{X}^m$ into the $d_m$-dimensional embedding representation $\mathbf{E}^m$, \ie, $\mathbf{E}^m = \mathcal{P}_{\theta^m}(\mathbf{X}^m) $. To overcome the view discrepancy, many MVL methods introduce view-level weights $\{w^m \}_{m=1}^V$ in their fusion models to achieve the balance across different views~\cite{zhang2024multimodal}, \eg, through weighted summation~\cite{zhou2020end,trostenMVC,zhang2023provable} or concatenation~\cite{abavisani2018deep,sun2020learning,xu2023self}.

Despite achieving some success, previous methods often infer the same weight for all $N$ samples in one view (\eg, $w^m$ for $\mathbf{X}^m$ and $w^n$ for $\mathbf{X}^n$).
This view-level weighting may not be suitable for every specific sample.
For example, the data of some samples in a low-quality view might be useful, but the model assigns a small weight to all samples in this view, resulting in the beneficial effects of these data being ignored during fusion.
To address this issue, we expect more fine-grained weighting strategies for multi-view fusion and propose the sample-level attention based multi-view fusion by improving Eq.~(\ref{Wfuison}). Specifically, for each multi-view data $\{\mathbf{x}_i^m\}_{m=1}^V$ from $\{\mathbf{X}^m\}_{m=1}^V$, we have the following paradigm
\begin{equation}\label{Sfuison}
\mathbf{z}_i= \mathcal{F}_{\theta_f}(\mathbf{A}_i, \{\mathbf{e}_i^m \}_{m=1}^V)= \mathcal{F}_{\theta_f}(\mathbf{A}_i,\{ \mathcal{P}_{\theta^m}(\mathbf{x}_i^m)\}_{m=1}^V),
\end{equation}
where $\mathbf{z}_i \in \mathbb{R}^{d}$, $\mathbf{A}_i \in \mathbb{R}^{V \times V}$, $\mathbf{e}_i^m \in \mathbb{R}^{d_m}$, $\mathbf{x}_i^m \in \mathbb{R}^{D_m}$.
To implement Eq.~(\ref{Sfuison}), we are motivated by the self-attention mechanism in sequence modeling and propose a multi-view transformer fusion network as shown in Figure~\ref{rml}(a).
Concretely, for the $i$-th input data, we treat its $V$ views $\{\mathbf{x}_i^1; \mathbf{x}_i^2; \dots; \mathbf{x}_i^V\}$ as $V$ words in a sentence, and then use multiple multilayer perceptrons (MLPs) to obtain multi-view word embeddings or called tokens~\cite{vaswani2017attention,kenton2019bert}, \ie, we implement $\{\mathbf{e}_i^m = {\rm MLP}_{\theta^m}(\mathbf{x}_i^m) \in \mathbb{R}^{d_e}\}_{m=1}^V$ which simultaneously unifies the heterogeneous data format of different views.
Then, for the multi-view sentence $\mathbf{E}_i = [\mathbf{e}_i^1; \mathbf{e}_i^2; \dots; \mathbf{e}_i^V] \in \mathbb{R}^{V \times d_e}$, we use trainable projection matrices $\mathbf{W}_q,\mathbf{W}_k,\mathbf{W}_v \in \mathbb{R}^{d_e \times d_e}$ to obtain the queries $\mathbf{Q}_i = \mathbf{E}_i \mathbf{W}_q$, keys $\mathbf{K}_i = \mathbf{E}_i \mathbf{W}_k $, and values $\mathbf{V}_i = \mathbf{E}_i \mathbf{W}_v$.
For the $i$-th input data, the attention scores among $V$ views are calculated by using the following formula:
\begin{equation}
    \mathbf{A}_i = {\rm{softmax}} \left( \mathbf{Q}_i  \mathbf{K}_i^T / \sqrt{d_e} \right) \in \mathbb{R}^{V \times V},
\end{equation}
where $\mathbf{A}_i$ is the sample-level attention score matrix among $V$ views of the $i$-th sample. The sample-level attention based representation of the $i$-th data is learned by
\begin{equation}
    \hat{\mathbf{E}}_i = \mathbf{A}_i \mathbf{V}_i = \mathbf{A}_i \mathbf{E}_i \mathbf{W}_v \in \mathbb{R}^{V \times d_e}.
\end{equation}
Upon $\hat{\mathbf{E}}_i$, we adopt the feed-forward neural network (FFN) as well as the residual connection in transformers~\cite{vaswani2017attention} to increase the representation capability of our model:
\begin{equation}
\begin{aligned}
    \mathbf{R}_i & = \hat{\mathbf{E}}_i + \mathbf{E}_i \in \mathbb{R}^{V \times d_e}, \\
    \mathbf{F}_i & = \mathbf{R}_i + {\rm FFN}_{\zeta}(\mathbf{R}_i) \in \mathbb{R}^{V \times d_e},
\end{aligned}
\end{equation}
where $\mathbf{R}_i$ and $\mathbf{F}_i$ denote the representations through the residual connection, $\mathbf{F}_i = [\mathbf{f}_i^1;\mathbf{f}_i^2;\dots;\mathbf{f}_i^V]$ is the encoded embeddings corresponding to $V$ views.
For the $i$-th data, we linearly add all encoded embeddings together and then utilize another MLP to obtain the fused representation $\mathbf{z}_i \in \mathbf{Z}$:
\begin{equation}
\begin{aligned}
    \mathbf{z}_i & = {\rm MLP}_{\phi}(\mathbf{f}_i^1+\mathbf{f}_i^2+\dots+\mathbf{f}_i^V) \in \mathbb{R}^{d}.
\end{aligned}
\end{equation}
Overall, the network parameters $\theta_f$ to be optimized in our implemented multi-view fusion model $\mathcal{F}_{\theta_f}$ include $\{\{\theta^m\}_{m=1}^V, \mathbf{W}_q,\mathbf{W}_k,\mathbf{W}_v, \zeta, \phi \}$.
For a sentence, it is established that transformers usually capture the information in semantic words while disregarding the empty words~\cite{vaswani2017attention,kenton2019bert,minaee2021deep}.
Similarly, our method treats different views of a multi-view sample as distinct words.
By employing the sample-level attention, our multi-view transformer fusion network capture the interrelationships within imperfect views and is expected to focus on the usable views, thus enabling effective multi-view representation fusion.

\subsection{Simulated perturbation based multi-view contrastive learning}
Multi-view contrastive learning is widely applied to achieve the information interaction among different views by aligning their representations.
However, directly forcing alignment among all views might lead to high-quality ones being negatively impacted by low-quality ones~\cite{Xu_2023_NIPS}.
Moreover, imperfect data in real-world applications is usual, and thus some collected views typically contain noise or unusable data.
In this paper, we will not pursue the alignment between views but propose a novel simulated perturbation based multi-view contrastive learning method, for training a robust multi-view fusion model.
As shown in Figure~\ref{rml}(b), we achieve multi-view information interaction by the representation alignment between two data simulated perturbations.

Concretely,
motivated by the success of data augmentation~\cite{van2001art,wang2022contrastive,Luo2024SCM,wang2025augrefer},
to enhance the model robustness to imperfect multiple views, we randomly add the noise perturbation and the unusable perturbation on partial views to respectively simulate the noisy data and unusable data.
The model is trained to achieve the representation alignment before and after dynamic perturbations for eventually resisting them.
Formally, we define the simulated perturbations as:
\begin{definition}[Noise Perturbation]
Given a multi-view dataset $\{{\mathbf X}^{1},{\mathbf X}^{2},\dots ,{\mathbf X}^{V}\}$, the simulated noise perturbation obtains
\begin{equation}
[{\mathbf N}^{1},{\mathbf N}^{2},\dots ,{\mathbf N}^{V}] = \mathcal{S}_{p, \sigma}^N({\mathbf X}^{1},{\mathbf X}^{2},\dots ,{\mathbf X}^{V}).
\end{equation}
The function $\mathcal{S}_{p, \sigma}^N$ performs the process of adding noise to the data.
In this process, for any ${\mathbf x}_i^m$ from ${\mathbf X}^m$, $i \in \{1,2,\dots,N\}, m \in \{1,2,\dots,V\}$, we compute the corresponding ${\mathbf n}_i^{m} \in {\mathbf N}^{m}$ by
\begin{equation}
\begin{aligned}
{\mathbf n}_i^{m} =
\begin{matrix}
\begin{cases}
{\mathbf x}^m_i + \epsilon_i^m & \emph{if}~~\delta_i^m < p, \\
{\mathbf x}^m_i                & \emph{else},
\end{cases}
\end{matrix}
\end{aligned}
\end{equation}
where $\epsilon_i^m \in \mathbb{R}^{D_m}$ and $\delta_i^m \in \mathbb{R}$ are randomly sampled from the Gaussian distribution $\mathcal{N}(0, \sigma^2)$ and uniform distribution $\mathcal{U}(0,1)$, respectively, i.e., $\epsilon_i^m \sim \mathcal{N}(0, \sigma^2)$, $\delta_i^m \sim \mathcal{U}(0,1)$. $0\leq p\leq 1$ controls the ratio of data perturbed by random noise to the overall data.
\end{definition}
\begin{definition}[Unusable Perturbation]
Given a multi-view dataset $\{{\mathbf X}^{1},{\mathbf X}^{2},\dots ,{\mathbf X}^{V}\}$, the simulated unusable perturbation obtains
\begin{equation}
[{\mathbf M}^{1},{\mathbf M}^{2},\dots ,{\mathbf M}^{V}] = \mathcal{S}_{r}^M({\mathbf X}^{1},{\mathbf X}^{2},\dots ,{\mathbf X}^{V}).
\end{equation}
The function $\mathcal{S}_{r}^M$ performs the process of dropping data.
In this process, we have a random indicator matrix $\mathbf{A} \in \{0,1\}^{N \times V}$ and for any ${\mathbf x}_i^m$ from ${\mathbf X}^m$, $a_{im}$ from $\mathbf{A}$, $i \in \{1,2,\dots,N\}, m \in \{1,2,\dots,V\}$, the corresponding ${\mathbf m}_i^{m} \in {\mathbf M}^{m}$ is generated by
\begin{equation}
\begin{aligned}
{\mathbf m}_i^{m} =
\begin{matrix}
\begin{cases}
{\mathbf x}^m_i & \emph{if}~~a_{im} = 1, \\
{\mathbf 0}     & \emph{else},
\end{cases}
\end{matrix}
\end{aligned}
\end{equation}
where ${\mathbf m}_i^{m} = {\mathbf 0}$ simulates that ${\mathbf x}^m_i$ becomes unusable data, while we have the constraint $\sum_{m=1}^V a_{im} > 0$ guaranteeing that at least one view data remains available for the $i$-th sample.
Letting $\mathbb{I}\{\cdot\}$ represent the indicator function, i.e., $\mathbb{I}\{{\rm True}\} = 1$; otherwise $\mathbb{I}\{{\rm False}\} = 0$, we have another constraint $\sum_{i} (\mathbb{I}\{\sum_m a_{im} < V\}) / N = r$, and $0\leq r \leq 1$ controls the ratio of unusable data to all available data.
\end{definition}
For the multi-view data processed through the above two simulated perturbations, we then leverage our proposed multi-view transformer fusion network $\mathcal{F}_{\theta_f}$ to learn the corresponding fused representations $\mathbf{Z}^N$ and $\mathbf{Z}^M$:
\begin{equation}
\begin{aligned}
\mathbf{Z}^N &= \mathcal{F}_{\theta_f}({\mathbf N}^{1},{\mathbf N}^{2},\dots ,{\mathbf N}^{V}), \\
\mathbf{Z}^M &= \mathcal{F}_{\theta_f}({\mathbf M}^{1},{\mathbf M}^{2},\dots ,{\mathbf M}^{V}).
\end{aligned}
\end{equation}
During model training, we randomly apply the simulated noise and unusable perturbations to partial views in multi-view data, resulting in dynamically changing $\mathbf{Z}^N$ and $\mathbf{Z}^M$. Contrastive learning between $\mathbf{Z}^N$ and $\mathbf{Z}^M$ is then performed to encourage their representation alignment as well as discrimination.
Specifically, our method optimizes the parameter $\theta_f$ in the multi-view fusion model $\mathcal{F}_{\theta_f}$ by minimizing the following loss function $\mathcal{L}_{\rm RML}$:
\begin{small}
\begin{equation}\label{con}
\begin{aligned}
    \mathcal{L}_{\rm RML} &= - \frac{1}{n} \sum_{i=1}^{n} \left [ \mathcal{L}_{\rm InfoNCE} (\mathbf{z}^{N}_i,\mathbf{Z}^{M}) + \mathcal{L}_{\rm InfoNCE} (\mathbf{z}^{M}_i,\mathbf{Z}^{N}) \right]  \\
                      &= - \frac{1}{n} \sum_{i=1}^{n} \log \frac{e^{d({\mathbf{z}}_i^N, {\mathbf{z}}_i^M)/\tau}}{e^{d({\mathbf{z}}_i^N, {\mathbf{z}}_i^M)/\tau} + \sum_{{\mathbf{z}} \in \mathcal{N}_i^{N}} e^{d({\mathbf{z}}_i^N, {\mathbf{z}})/\tau}} \\
                      &~~~~- \frac{1}{n} \sum_{i=1}^{n} \log \frac{e^{d({\mathbf{z}}_i^M, {\mathbf{z}}_i^N)/\tau}}{e^{d({\mathbf{z}}_i^M, {\mathbf{z}}_i^N)/\tau} + \sum_{{\mathbf{z}} \in \mathcal{N}_i^{M}} e^{d({\mathbf{z}}_i^M, {\mathbf{z}})/\tau}}, 
\end{aligned}
\end{equation}
\end{small}where $d({\mathbf{z}}_i^N, {\mathbf{z}}_i^M) = \frac{{\mathbf{z}}_i^N \cdot {\mathbf{z}}_i^M}{\|{\mathbf{z}}_i^N\|_2\|{\mathbf{z}}_i^M\|_2}$ measures the cosine similarity between two sample representations. $\tau$ is a controllable temperature parameter in the InfoNCE loss.
For $\mathbf{z}_i^N$, $\mathcal{N}_i^{N} = \{{\mathbf{z}}_j^v\}_{j \neq i}^{v=N,M}$ denotes the set of representations to construct the negative sample pairs, \ie, $\{{\mathbf{z}}_i^N, {\mathbf{z}}_j^v\}_{j \neq i}^{v=N,M}$.
Similarly, for $\mathbf{z}_i^M$, $\mathcal{N}_i^{M} = \{{\mathbf{z}}_j^v\}_{j \neq i}^{v=N,M}$ and the negative sample pairs are $\{{\mathbf{z}}_i^M, {\mathbf{z}}_j^v\}_{j \neq i}^{v=N,M}$.
By minimizing Eq.~(\ref{con}), the multi-view contrastive learning brings similar fused representations closer and pushes dissimilar ones apart.
This facilitates the model to capture discriminative information across multi-view data for benefiting downstream tasks.

Unlike previous methods, our method utilizes the multi-view transformer fusion network to achieve the sample-level attention based representation fusion for imperfect multi-view data.
On the fused representations, our method further leverages the simulated perturbation based multi-view contrastive learning to perform representation alignment.
These two components are integrated into a unified framework which encourages the model to focus on the clean views among partially noisy multi-view data, and to focus on the usable views among partially unusable multi-view data.
This not only enhances the model robustness to low-quality views, but also promotes the extraction of useful discriminative information among high-quality views.

\subsection{Multi-view learning with RML regularization}

\begin{table*}[!ht]
\caption{Performance comparison on unsupervised multi-view clustering (Bold indicates the latest best results)}\label{tableall}
\vspace{-0.2cm}
\centering
\resizebox{\textwidth}{!}{
\begin{threeparttable}
    \begin{tabular}{lcccccccccccccccccc}
    \toprule
    \multirow{1}{*}{Method} 
    &\multicolumn{2}{c}{DHA} &\multicolumn{2}{c}{BDGP} &\multicolumn{2}{c}{Prokaryotic} &\multicolumn{2}{c}{Cora} &\multicolumn{2}{c}{YoutubeVideo} &\multicolumn{2}{c}{WebKB} &\multicolumn{2}{c}{VOC} &\multicolumn{2}{c}{NGs} &\multicolumn{2}{c}{Cifar100} \\
    \cmidrule(r){2-3} \cmidrule(r){4-5} \cmidrule(r){6-7} \cmidrule(r){8-9} \cmidrule(r){10-11} \cmidrule(r){12-13} \cmidrule(r){14-15} \cmidrule(r){16-17} \cmidrule(r){18-19}
    & ACC & NMI & ACC & NMI & ACC & NMI & ACC & NMI & ACC & NMI & ACC & NMI & ACC & NMI & ACC & NMI & ACC & NMI \\
    \hline
    K-means~\cite{macqueen1967some}   &0.656 &0.798 &0.443 &0.573 &0.562 &\textbf{0.325}  &0.363 &0.172 &0.199 &0.194 &0.617 &0.002 &0.487 &0.360 &0.206 &0.019 &0.975 &0.996 \\

    MCN~\cite{chen2021multimodal}     &0.758 &0.800 &0.957 &0.901 &0.528 &0.287 &0.386 &0.184 &0.183 &0.187 &0.636 &0.081 &0.274 &0.286 &0.886 &0.736 &0.864 &0.962 \\
    
    CPSPAN~\cite{jin2023deep}         &0.663 &0.775 &0.690 &0.636 &0.539 &0.229  &0.419 &0.190 &0.232 &0.221 &0.771 &0.166 &0.452 &0.488 &0.352 &0.215 &0.918 &0.982 \\
    CVCL~\cite{chen2023deep}          &0.662 &0.754 &0.907 &0.785 &0.526 &0.281  &0.483 &0.310 &0.273 &0.258 &0.741 &0.246 &0.315 &0.317 &0.568 &0.317 &0.956 &0.977 \\
    DSIMVC~\cite{tang2022deepi}       &0.635 &0.778 &0.983 &0.944 &0.597 &0.318  &0.478 &0.353 &0.189 &0.188 &0.702 &0.250 &0.212 &0.204 &0.630 &0.502 &0.895 &0.969 \\
    DSMVC~\cite{tang2022deep}         &0.762 &0.836 &0.523 &0.396 &0.502 &0.258  &0.447 &0.308 &0.178 &0.180 &0.663 &0.134 &0.633 &\textbf{0.723} &0.352 &0.082 &0.851 &0.959 \\
    MFLVC~\cite{xu2022multi}          &0.716 &0.812 &\textbf{0.983} &\textbf{0.951} &0.569 &0.316  &0.485 &0.351 &0.184 &0.186 &0.672 &0.245 &0.292 &0.280 &0.908 &0.802 &0.877 &0.964 \\
    SCM~\cite{Luo2024SCM}             &0.814 &0.840 &0.962 &0.885 &0.550 &0.278  &0.564 &\textbf{0.378} &0.316 &0.313 &0.689 &0.094 &0.607 &0.622 &0.968 &0.900 &0.999 &0.999 \\
    SCM$_{RE}$~\cite{Luo2024SCM}      &0.804 &0.840 &0.971 &0.913 &0.582 &0.312  &\textbf{0.574} &0.374 &0.317 &0.322 &0.725 &0.268 &0.629 &0.629 &0.965 &0.893 &0.999 &0.999 \\
    \hline
    RML+K-means     &\textbf{0.822} &\textbf{0.847} &0.981 &0.941 &\textbf{0.605} &0.316  &0.570 &0.371 &\textbf{0.331} &\textbf{0.339} &\textbf{0.868} &\textbf{0.508} &\textbf{0.656} &0.615 &\textbf{0.983} &\textbf{0.943} &\textbf{0.999} &\textbf{0.999} \\
    \bottomrule
    \end{tabular}
\end{threeparttable}}
\vspace{-0.2cm}
\end{table*}

\begin{table}[!t]
\caption{Descriptions of benchmark datasets used in this paper}\label{table0}
\vspace{-0.2cm}
\centering
\resizebox{0.49\textwidth}{!}{
\begin{threeparttable}
    \begin{tabular}{lccccc}
    \toprule
    Name           & Type  & Features & \#Sample & \#Class \\
    \hline
    DHA~\cite{lin2012human}                & human motions & 110 - 6144         & 483     & 23 \\
    BDGP~\cite{cai2012joint}               & drosophila embryos & 1750 - 79          & 2,500   & 5 \\
    Prokaryotic~\cite{brbic2016landscape}  & prokaryotic species & 393 - 3 - 438      & 551     & 4 \\
    Cora~\cite{bisson2012co}               & scientific documents & 2708 - 1433        & 2,708   & 7 \\
    YoutubeVideo~\cite{madani2013using}    & video data & 512 - 647 - 838    & 101,499 & 31 \\
    WebKB~\cite{sun2007kernelized}         & web pages  & 2949 - 334         & 1,051   & 2 \\
    VOC~\cite{everingham2010pascal}        & image-text pairs  & 512 - 399          & 5,649   & 20 \\
    NGs~\cite{hussain2010improved}         & multi-features of news  & 2000 - 2000 - 2000 & 500     & 5 \\
    Cifar100~\cite{krizhevsky2009learning} & multi-features of images  & 512 - 1024 - 2048  & 50,000  & 100 \\
    MIRFLICKR-25K~\cite{huiskes2008mir}    & image-text pairs & 4096 - 1386        & 20,015  & 24 \\
    NUS-WIDE~\cite{rasiwasia2010new}       & image-text pairs & 4096 - 1000        & 186,577 & 10 \\
    \bottomrule
    \end{tabular}
\end{threeparttable}
}
\vspace{-0.2cm}
\end{table}

Our proposed robust multi-view learning method RML can not only perform multi-view representation learning in a self-supervised manner, but can also be used as a plug-and-play regularization to enhance other multi-view methods shown in Figure~\ref{rml}(c).
Next, we formulate the multi-view learning methods and demonstrate how to integrate RML into them.

Considering that different multi-view methods have different tasks and adopt inconsistent model structures for handling multi-view data of different application domains, we first decompose multi-view methods into the representation learning module $\mathcal{R}_{\theta_l}$ and the task-specific module $\mathcal{T}_{\theta_t}$, where the parameters to be optimized are distinguished by $\theta_l$ and $\theta_t$, respectively.
Then, we formally decompose the framework of multi-view learning methods as follows:
\begin{small}
\begin{equation}\label{task}
\begin{aligned}
\mathcal{L}_{\rm Task} &:= {\rm Loss}_{\rm task}(\mathcal{T}_{\theta_t}(\{{\mathbf H}^{m}\}_{m=1}^V), \mathcal{P}) \\
&~\mathrm{s.t.}~[{\mathbf H}^{1},{\mathbf H}^{2},\dots ,{\mathbf H}^{V}] = \mathcal{R}_{\theta_l}({\mathbf X}^{1},{\mathbf X}^{2},\dots ,{\mathbf X}^{V}),
\end{aligned}
\end{equation}
\end{small}where $[{\mathbf H}^{1},{\mathbf H}^{2},\dots ,{\mathbf H}^{V}]$ denotes the multi-view hidden representations learned by the module $\mathcal{R}_{\theta_l}$, $\mathcal{L}_{\rm Task}$ is the task-specific loss function defined upon the hidden representations through the task-specific module $\mathcal{T}_{\theta_t}$ together with the extra task-specific supervision signals $\mathcal{P}$, \eg, the cross entropy loss and sample labels.
In this way, we do not need to modify the details of how specific methods handle multi-view data, and our RML can be easily integrated into Eq.~(\ref{task}) as a regularization term for joint optimization:
\begin{small}
\begin{equation}\label{RMLtask}
\begin{aligned}
\mathcal{L} &= \mathcal{L}_{\rm Task} (\mathcal{T}_{\theta_t}(\{{\mathbf H}^{m}\}_{m=1}^V), \mathcal{P}) + \lambda \mathcal{L}_{\rm RML} (\mathbf{Z}^N, \mathbf{Z}^M) \\
&~\mathrm{s.t.}~[{\mathbf H}^{1},{\mathbf H}^{2},\dots ,{\mathbf H}^{V}] = \mathcal{R}_{\theta_l}({\mathbf X}^{1},{\mathbf X}^{2},\dots ,{\mathbf X}^{V}), \\
&~~~~~~~~[{\mathbf N}^{1},{\mathbf N}^{2},\dots ,{\mathbf N}^{V}] = \mathcal{S}_{p, \sigma}^N({\mathbf H}^{1},{\mathbf H}^{2},\dots ,{\mathbf H}^{V}), \\
&~~~~~~~~[{\mathbf M}^{1},{\mathbf M}^{2},\dots ,{\mathbf M}^{V}] = \mathcal{S}_{r}^M({\mathbf H}^{1},{\mathbf H}^{2},\dots ,{\mathbf H}^{V}), \\
&~~~~~~~~\mathbf{Z}^N = \mathcal{F}_{\theta_f}({\mathbf N}^{1},{\mathbf N}^{2},\dots ,{\mathbf N}^{V}), \\
&~~~~~~~~\mathbf{Z}^M = \mathcal{F}_{\theta_f}({\mathbf M}^{1},{\mathbf M}^{2},\dots ,{\mathbf M}^{V}).
\end{aligned}
\end{equation}
\end{small}When minimizing $\mathcal{L}_{\rm Task} (\mathcal{T}_{\theta_t}(\{{\mathbf H}^{m}\}_{m=1}^V), \mathcal{P})$ in other multi-view methods, we actually treat the multi-view representations $[{\mathbf H}^{1},{\mathbf H}^{2},\dots ,{\mathbf H}^{V}]$ as the input of our multi-view transformer fusion network $\mathcal{F}_{\theta_f}$, and leverage the loss of our simulated perturbation based multi-view contrastive learning $\mathcal{L}_{\rm RML} (\mathbf{Z}^N, \mathbf{Z}^M)$ to regularize the representation learning.

Finally, for $n$ samples in a multi-view dataset ($n$ is the batch size), we formulate the mini-batch update rules of parameters in the
 representation learning module, task-specific module, and RML regularization module as follows:
\begin{equation}
\begin{aligned}
\begin{matrix}
\begin{cases}
\theta_l \leftarrow \theta_l - \frac{\eta}{n} \sum_{i=1}^{n} \left (\frac{\partial \mathcal{L}_{\rm Task}}{\partial \theta_l} + \lambda \frac{\partial \mathcal{L}_{\rm RML}}{\partial \theta_l} \right) \\
\theta_t \leftarrow \theta_t - \frac{\eta}{n} \sum_{i=1}^{n} \frac{\partial \mathcal{L}_{\rm Task}}{\partial \theta_t} \\
\theta_f \leftarrow \theta_f - \frac{\eta}{n} \sum_{i=1}^{n} \lambda \frac{\partial \mathcal{L}_{\rm RML}}{\partial \theta_f}
\end{cases}
\end{matrix}
\end{aligned}
\end{equation}
where $\eta$ is the learning rate and $\lambda$ is the trade-off between $\mathcal{L}_{\rm RML}$ and $\mathcal{L}_{\rm Task}$. In this way, our proposed regularization term influences the parameter $\theta_l$ of the representation learning module $\mathcal{R}_{\theta_l}$ in the specific multi-view method and the parameter $\theta_f$ in our RML module $\mathcal{F}_{\theta_f}$.
This is expected to make the overall model learn more robust and discriminative representations, thereby promoting specific multi-view tasks.
We will experimentally validate this in the next section.

\section{Main Results}\label{EX}

\textbf{Datasets.}
Multi-view data is prevalent in real-world applications and exhibits significant heterogeneity.
Different datasets usually vary in data modalities, dimensions, sparsity, the number and format of views.
Since a MVL method which is compatible with various multi-view datasets is highly anticipated, we conducted experiments on multiple types of multi-view datasets to validate the effectiveness and universality of our method. The information of the used benchmark datasets is listed in Table~\ref{table0} showing considerable diversity\footnote{For more detailed dataset information, related work, experimental settings, and comparison results, please refer to the Appendix or our Code.}.

\subsection{RML on unsupervised multi-view clustering}
\textbf{Settings.}
In this part, we conduct representation learning and clustering by RML to evaluate its performance of multi-view fusion.
To be specific, we first utilize RML to learn fused representations on extensive multi-view datasets (\ie, DHA, BDGP, Prokaryotic, Cora, YoutubeVideo, WebKB, VOC, NGs, and Cifar100), and then perform K-Means on the fused representations to show the clustering quality of RML.
The comparison methods include the classical K-Means~\cite{macqueen1967some} and eight recent deep-learning-based multi-view clustering methods:
{MCN}~\cite{chen2021multimodal},
{CPSPAN}~\cite{jin2023deep},
{CVCL}~\cite{chen2023deep},
{DSIMVC}~\cite{tang2022deepi},
{DSMVC}~\cite{tang2022deep},
{MFLVC}~\cite{xu2022multi}, {SCM}~\cite{Luo2024SCM}, and {SCM$_{RE}$}~\cite{Luo2024SCM}.

\begin{table*}[!ht]
\caption{Performance comparison on noise-label multi-view classification with 50\% noise label rate (Trans. denotes the transformer network)}\label{table2p}
\vspace{-0.2cm}
\centering
\resizebox{\textwidth}{!}{
\begin{threeparttable}
    \begin{tabular}{lccccccccccccccc}
    \toprule
    \multirow{1}{*}{Method} 
    &\multicolumn{3}{c}{DHA} &\multicolumn{3}{c}{BDGP} &\multicolumn{3}{c}{Prokaryotic} &\multicolumn{3}{c}{Cora} &\multicolumn{3}{c}{YoutubeVideo} \\
    \cmidrule(r){2-4} \cmidrule(r){5-7} \cmidrule(r){8-10} \cmidrule(r){11-13} \cmidrule(r){14-16}
    & ACC & Pre. & F1 & ACC & Pre. & F1 & ACC & Pre. & F1 & ACC & Pre. & F1 & ACC & Pre. & F1 \\
    \hline

    Trans.+$\mathcal{L}_{\rm CE}$      &0.457 &0.487 &0.448 &0.437 &0.441 &0.435 &0.473 &0.594 &0.505 &0.400 &0.432 &0.407 &0.266 &0.804 &0.112 \\
    \rowcolor{gray!10}
    Trans.+$\mathcal{L}_{\rm MCE}$     &0.470 &0.519 &0.467 &0.442 &0.446 &0.441 &0.472 &0.606 &0.505 &0.374 &0.413 &0.382 &0.267 &0.805 &0.112 \\
    RML+$\mathcal{L}_{\rm CE}$         &0.608 &0.736 &0.563 &0.933 &0.933 &0.933 &0.735 &0.783 &0.747 &0.664 &0.669 &0.648 &0.592 &0.634 &0.584 \\
    \rowcolor{gray!10}
    RML+$\mathcal{L}_{\rm MCE}$        &0.610 &0.737 &0.565 &0.936 &0.936 &0.936 &0.735 &0.783 &0.747 &0.665 &0.666 &0.651 &0.598 &0.639 &0.593 \\

    \bottomrule
    \end{tabular}
\end{threeparttable}}
\vspace{-0.2cm}
\end{table*}

\noindent\textbf{Comparison experiments.}
The performance is evaluated by the commonly-used metrics including clustering accuracy (ACC) and normalized mutual information (NMI), and we report average results of 5 independent runs as shown in Table~\ref{tableall}.
From the experimental results, we observe that:
I) A single method might be unable to perform well across different datasets. For example, method DSMVC performs well on DHA and VOC, but poorly on BDGP and NGs. Conversely, method MFLVC has good performance on BDGP and NGs, but is less effective on DHA and VOC. This is due to the heterogeneity among various multi-view datasets, which makes it difficult for specific methods to effectively address all multi-view scenarios. Therefore, ensuring that multi-view learning methods are as compatible as possible with a wider variety of datasets is a crucial research goal.
II) Our RML achieved the best or comparable performance. For instance, on datasets DHA, YoutubeVideo, WebKB, VOC, NGs, and Cifar100, our method outperformed the best comparison methods. On datasets BDGP, Prokaryotic, and Cora, our RML also approached the performance of the best methods. These results indicate that our RML can effectively perform unsupervised representation learning and information fusion on diverse datasets.
This can be attributed to our proposed novel multi-view transformer fusion network and simulated perturbation based contrastive learning strategy.

\begin{table*}[!ht]
\caption{Performance comparison on cross-modal hashing retrieval (16, 32, 64, and 128 represent the different settings of hash code length)}\label{table3p}
\vspace{-0.2cm}
\centering
\resizebox{\textwidth}{!}{
\begin{threeparttable}
    \begin{tabular}{lcccccccccccccccc}
    \toprule
    \multirow{1}{*}{Method} &\multicolumn{8}{c}{MIRFLICKR-25K} &\multicolumn{8}{c}{NUS-WIDE} \cr
    \cmidrule(r){2-9} \cmidrule(r){10-17}
    &\multicolumn{4}{c}{Image $\rightarrow$ Text} &\multicolumn{4}{c}{Text $\rightarrow$ Image} &\multicolumn{4}{c}{Image $\rightarrow$ Text} &\multicolumn{4}{c}{Text $\rightarrow$ Image} \\
    \cmidrule(r){2-5} \cmidrule(r){6-9} \cmidrule(r){10-13} \cmidrule(r){14-17}
    & 16 & 32 & 64 & 128 & 16 & 32 & 64 & 128 & 16 & 32 & 64 & 128 & 16 & 32 & 64 & 128 \\
    \hline
    CVH~\cite{KumarU11}                &0.620 &0.608 &0.594 &0.583 &0.629 &0.615 &0.599 &0.587 &0.487 &0.495 &0.456 &0.419 &0.470 &0.475 &0.444 &0.412 \\
    FSH~\cite{LiuJWHZ17}               &0.581 &0.612 &0.635 &0.662 &0.576 &0.607 &0.635 &0.660 &0.557 &0.565 &0.598 &0.635 &0.569 &0.604 &0.651 &0.666 \\
    UGACH~\cite{ZhouDG14}              &0.685 &0.693 &0.704 &0.702 &0.673 &0.676 &0.686 &0.690 &0.613 &0.623 &0.628 &0.631 &0.603 &0.614 &0.640 &0.641 \\
    DJSRH~\cite{DingGZG16}             &0.652 &0.697 &0.700 &0.716 &0.662 &0.691 &0.683 &0.695 &0.502 &0.538 &0.527 &0.556 &0.465 &0.532 &0.538 &0.545 \\
    JDSH~\cite{ZhangPY18}              &0.724 &0.734 &0.741 &0.745 &0.710 &0.720 &0.733 &0.720 &0.647 &0.656 &0.679 &0.680 &0.649 &0.669 &0.689 &0.699 \\
    DGCPN~\cite{SuZZ19}                &0.711 &0.723 &0.737 &0.748 &0.695 &0.707 &0.725 &0.731 &0.610 &0.614 &0.635 &0.641 &0.617 &0.621 &0.642 &0.647 \\

    \hline
    
    UCCH~\cite{hu2022unsupervised}     &0.739 &0.744 &0.754 &0.760 &\textbf{0.725} &0.725 &0.743 &0.747 &0.698 &0.708 &0.737 &0.742 &0.701 &0.724 &0.745 &0.750 \\
    RML+UCCH &\textbf{0.745} &\textbf{0.763} &\textbf{0.769} &\textbf{0.769} &0.721 &\textbf{0.738} &\textbf{0.744} &\textbf{0.748} &\textbf{0.733} &\textbf{0.741} &\textbf{0.745} &\textbf{0.749} &\textbf{0.726} &\textbf{0.741} &\textbf{0.745} &\textbf{0.752} \\

    \hline
    NRCH~\cite{wang2024robust}         &0.760 &0.788 &0.785 &0.791 &0.747 &0.778 &0.780 &0.784  &0.627 &0.646 &\textbf{0.675} &0.670 &0.625 &\textbf{0.648} &\textbf{0.678} &0.665 \\
    RML+NRCH &\textbf{0.778} &\textbf{0.798} &\textbf{0.791} &\textbf{0.797} &\textbf{0.766} &\textbf{0.781} &\textbf{0.783} &\textbf{0.786}  &\textbf{0.660} &\textbf{0.653} &0.660 &\textbf{0.682} &\textbf{0.663} &0.640 &0.651 &\textbf{0.677} \\ 
    
    \bottomrule
    \end{tabular}

    \begin{tablenotes}
        \footnotesize
        \item[*]In this table, we compare RML+NRCH with NRCH and they are not compared with other cross-modal hashing retrieval methods due to the differences in their experimental settings.
    \end{tablenotes}
    
\end{threeparttable}}
\vspace{-0.2cm}
\end{table*}

\subsection{RML on noise-label multi-view classification}

\textbf{Settings.}
In this part, we conduct noise-label multi-view classification to evaluate the robustness of RML against noise labels.
Specifically, our experiments are carried out on datasets DHA, BDGP, Prokaryotic, Cora, and YoutubeVideo.
We follow the setting of noise-label learning \cite{HanYYNXHTS18,lukov2022teaching} and adopt the symmetric noise labels, which constructs the noise labels for a percentage of training samples by randomly replacing their truth labels with all possible labels.
The partition of training set and test set is $7:3$.
Our method includes $\rm RML+\mathcal{L}_{\rm CE}$ and $\rm RML+\mathcal{L}_{\rm MCE}$, 
which optimize the original cross-entropy loss $\mathcal{L}_{\rm CE}$ and multiple cross-entropy losses $\mathcal{L}_{\rm MCE}$ defined in Appendix, respectively.
Two baselines $\rm Trans.+\mathcal{L}_{CE}$ and $\rm Trans.+\mathcal{L}_{MCE}$ only minimize $\mathcal{L}_{\rm CE}$ and $\mathcal{L}_{\rm MCE}$, respectively, in which we adopt the same multi-view transformer fusion network as RML for a fair comparison.

\noindent\textbf{Comparison experiments.}
The performance is evaluated by the commonly-used metrics including classification accuracy (ACC), Precision (Pre.), and F1-score (F1). We report average results of 5 independent runs in Table~\ref{table2p} and have observations as follows:
I)
From the overall results, we observe that $\rm RML+\mathcal{L}_{\rm MCE}$ and $\rm Trans.+\mathcal{L}_{\rm MCE}$ respectively have better classification results than $\rm RML+\mathcal{L}_{\rm CE}$ and $\rm Trans.+\mathcal{L}_{\rm CE}$ in most cases.
Our proposed two simulated perturbations allow us to design new cross-entropy loss $\mathcal{L}_{\rm MCE}$, and optimizing it makes the model more effective to access the category information in imperfect multi-view data.
II)
Our proposed RML plays a positive role in the model robustness against noise labels.
The results show that the classification performance of $\rm RML+\mathcal{L}_{\rm MCE}$ and $\rm RML+\mathcal{L}_{\rm CE}$ consistently outperforms that of $\rm Trans.+\mathcal{L}_{\rm MCE}$ and $\rm Trans.+\mathcal{L}_{\rm CE}$.
For example, when the noise label rate is $50\%$, $\rm Trans.+\mathcal{L}_{\rm CE}$ has the accuracy of only $0.437$ on BDGP and $\rm RML+\mathcal{L}_{\rm MCE}$ improves it to $0.936$.
This suggests that our proposed $\mathcal{L}_{\rm RML}$ can be used as a superior regularization term for multi-view classification tasks, and the special designs in our framework promote the model's robustness against noisy labels.

\subsection{RML on cross-modal hashing retrieval}

\textbf{Settings.}
In this part, we conduct cross-modal hashing retrieval tasks to evaluate the effectiveness of RML as a plug-and-play approach.
Specifically, we use two image-text retrieval datasets (\ie, MIRFLICKR-25K and NUS-WIDE) and conduct two kinds of cross-modal retrieval task, \ie, $\rm Image \rightarrow Text$ utilizes an image query to retrieve the relevant text samples, $\rm Text \rightarrow Image$ uses a text query to retrieve the relevant image samples.
The comparison baselines of cross-modal hashing models includes two traditional methods
(CVH~\cite{KumarU11} and FSH~\cite{LiuJWHZ17})
and six deep methods (UGACH~\cite{ZhouDG14}, DJSRH~\cite{DingGZG16}, JDSH~\cite{ZhangPY18}, DGCPN~\cite{SuZZ19}, UCCH~\cite{hu2022unsupervised}, and NRCH~\cite{wang2024robust}). Two recent methods employ our proposed RML regularization and they are denoted as RML+UCCH and RML+NRCH, respectively.

\noindent\textbf{Comparison experiments.}
The retrieval results between images and texts are listed in Table~\ref{table3p}. The performance is evaluated by MAP Score and the larger is the better.
From the experimental results, we could obtain the following conclusions:
I) Deep learning based models often achieve better performance than traditional shallow models. For example, on MIRFLICKR-25K, deep methods (\ie, JDSH, DGCPN, UCCH, NRCH, our RML+UCCH and RML+NRCH) can reach a MAP score of over $0.70$, which is superior than around $0.60$ for shallow methods (\ie, CVH and FSH). Additionally, longer hash codes usually are more beneficial for retrieval tasks.
II) Our proposed method achieved the best overall results in cross-modal retrieval tasks. For instance, when the hash code length is set to $16$ on NUS-WIDE, our RML+UCCH outperformed UCCH by $3\%$ MAP score in the image-to-text retrieval task, and our RML+UCCH outperformed UCCH by $2\%$ MAP score in the text-to-image retrieval task.
The consistent improvements can also be observed between RML+NRCH and NRCH.
These results demonstrate that our RML can be successfully applied in cross-modal hashing retrieval, and validate its effectiveness as a regularization module to promote existing methods.

\section{Ablation Study and Analysis}

\begin{figure}[!t]
\centering
\begin{subfigure}{0.49\linewidth}
\includegraphics[width=\linewidth]{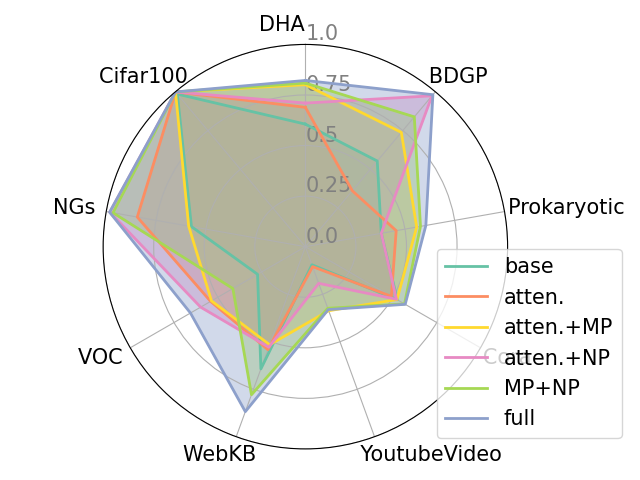}
\caption{ACC}
\vspace{-0.2cm}
\end{subfigure}
\begin{subfigure}{0.49\linewidth}
\includegraphics[width=\linewidth]{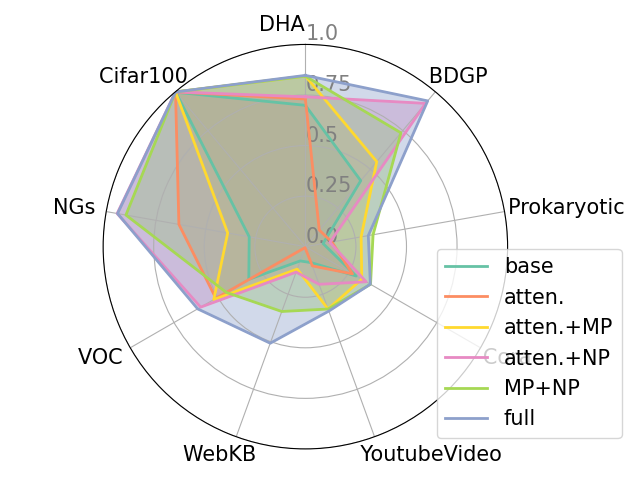}
\caption{NMI}
\vspace{-0.2cm}
\end{subfigure}
\caption{Ablation experiments of multi-view clustering.}\label{abcp}
\vspace{-0.2cm}
\end{figure}

\begin{table*}[!ht]
\caption{Ablation experiments of noise-label multi-view classification with RML+$\mathcal{L}_{\rm MCE}$ (the noise label rate is set to 50\%)}\label{table5p}
\vspace{-0.2cm}
\centering
\resizebox{\textwidth}{!}{
\begin{threeparttable}
    \begin{tabular}{lccccccccccccccc}
    \toprule
    \multirow{1}{*}{} 
    &\multicolumn{3}{c}{DHA} &\multicolumn{3}{c}{BDGP} &\multicolumn{3}{c}{Prokaryotic} &\multicolumn{3}{c}{Cora} &\multicolumn{3}{c}{YoutubeVideo} \\
    \cmidrule(r){2-4} \cmidrule(r){5-7} \cmidrule(r){8-10} \cmidrule(r){11-13} \cmidrule(r){14-16}
    & ACC & Pre. & F1 & ACC & Pre. & F1 & ACC & Pre. & F1 & ACC & Pre. & F1 & ACC & Pre. & F1 \\
    \hline
    base         &0.583 &0.723 &0.542 &0.864 &0.872 &0.864 &0.753 &0.801 &0.763 &0.617 &0.621 &0.603 &0.307 &0.347 &0.223 \\
    atten.       &0.551 &0.687 &0.497 &0.698 &0.700 &0.695 &0.715 &0.769 &0.726 &0.645 &0.646 &0.632 &0.337 &0.487 &0.262 \\
    atten.+MP~~~~~~~    &0.606 &0.738 &0.567 &0.834 &0.834 &0.832 &0.734 &0.772 &0.743 &0.646 &0.643 &0.631 &0.548 &0.579 &0.541 \\
    atten.+NP    &0.532 &0.670 &0.478 &0.957 &0.958 &0.957 &0.731 &0.776 &0.740 &0.645 &0.663 &0.631 &0.421 &0.536 &0.374 \\
    NP+MP        &0.678 &0.788 &0.633 &0.943 &0.944 &0.943 &0.706 &0.767 &0.714 &0.644 &0.662 &0.630 &0.581 &0.638 &0.568 \\
    full         &0.610 &0.737 &0.565 &0.936 &0.936 &0.936 &0.735 &0.783 &0.747 &0.665 &0.666 &0.651 &0.598 &0.639 &0.593 \\
    \bottomrule
    \end{tabular}
\end{threeparttable}}
\vspace{-0.2cm}
\end{table*}

\begin{figure*}[!ht]
\centering
\begin{subfigure}{0.49\linewidth}
\includegraphics[width=\linewidth]{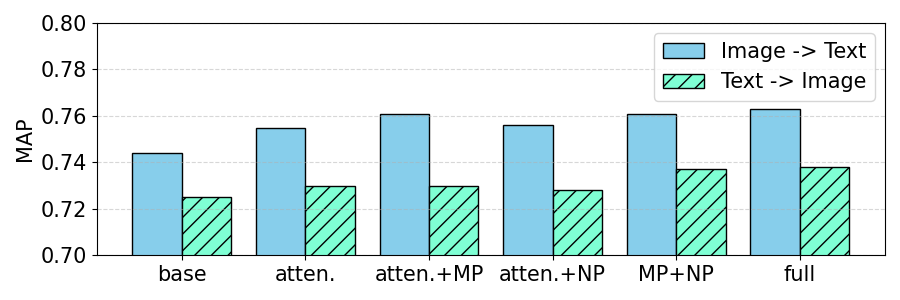}
\caption{MIRFLICKR-25K}
\vspace{-0.2cm}
\end{subfigure}~
\begin{subfigure}{0.49\linewidth}
\includegraphics[width=\linewidth]{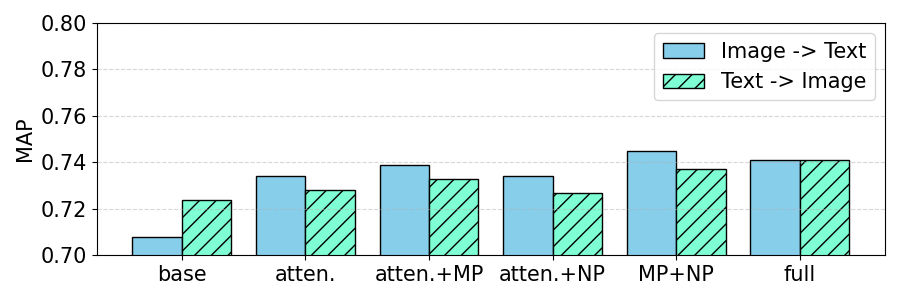}
\caption{NUS-WIDE}
\vspace{-0.2cm}
\end{subfigure}
\caption{Ablation experiments of cross-modal hashing retrieval with RML+UCCH on (a) MIRFLICKR-25K and (b) NUS-WIDE.}\label{abrp}
\vspace{-0.3cm}
\end{figure*}

\begin{figure}[!t]
\centering
\begin{subfigure}{0.49\linewidth}
\includegraphics[width=\linewidth]{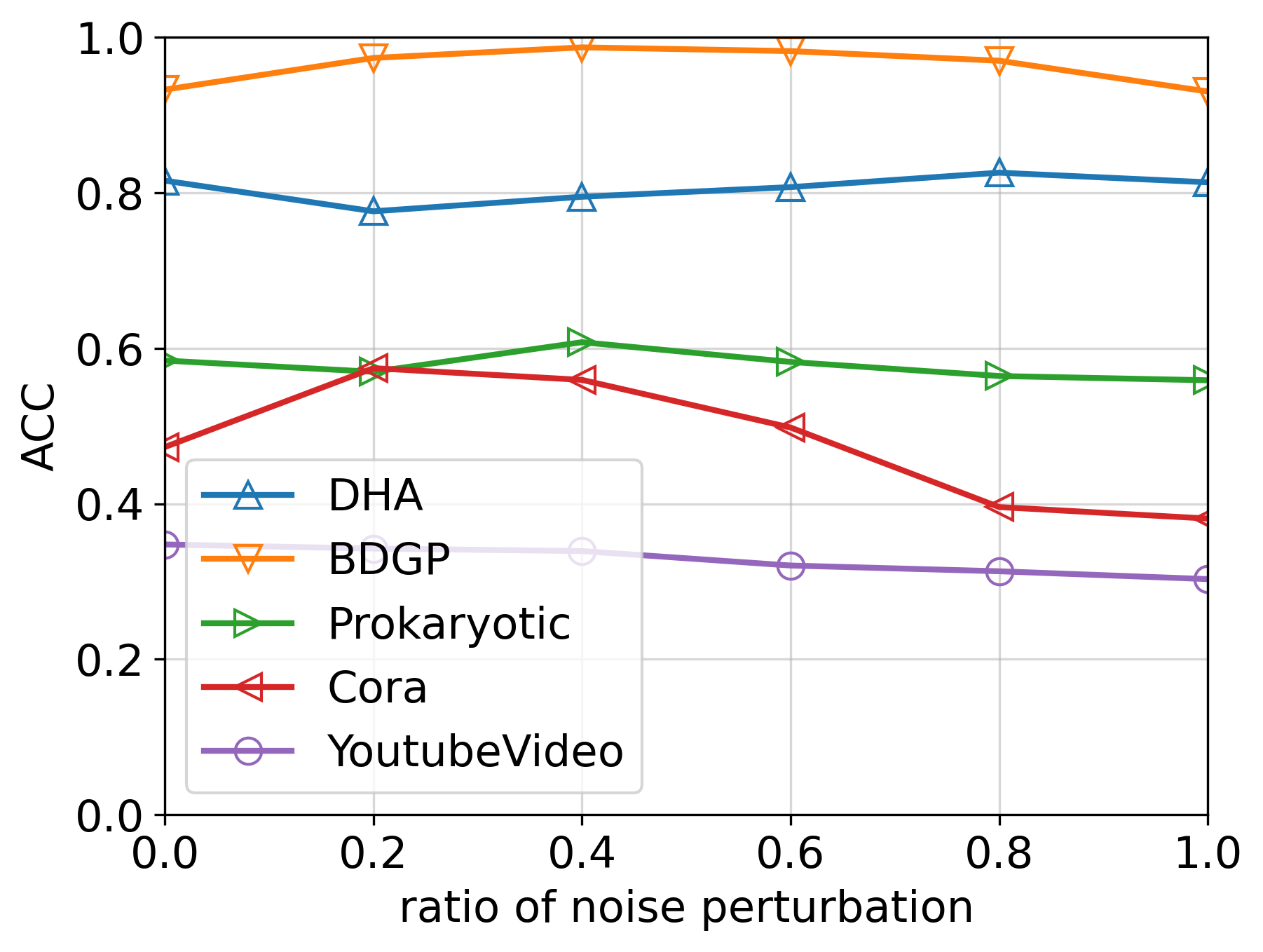}
\caption{$p$ $vs.$ ACC}
\vspace{-0.2cm}
\end{subfigure}
\begin{subfigure}{0.49\linewidth}
\includegraphics[width=\linewidth]{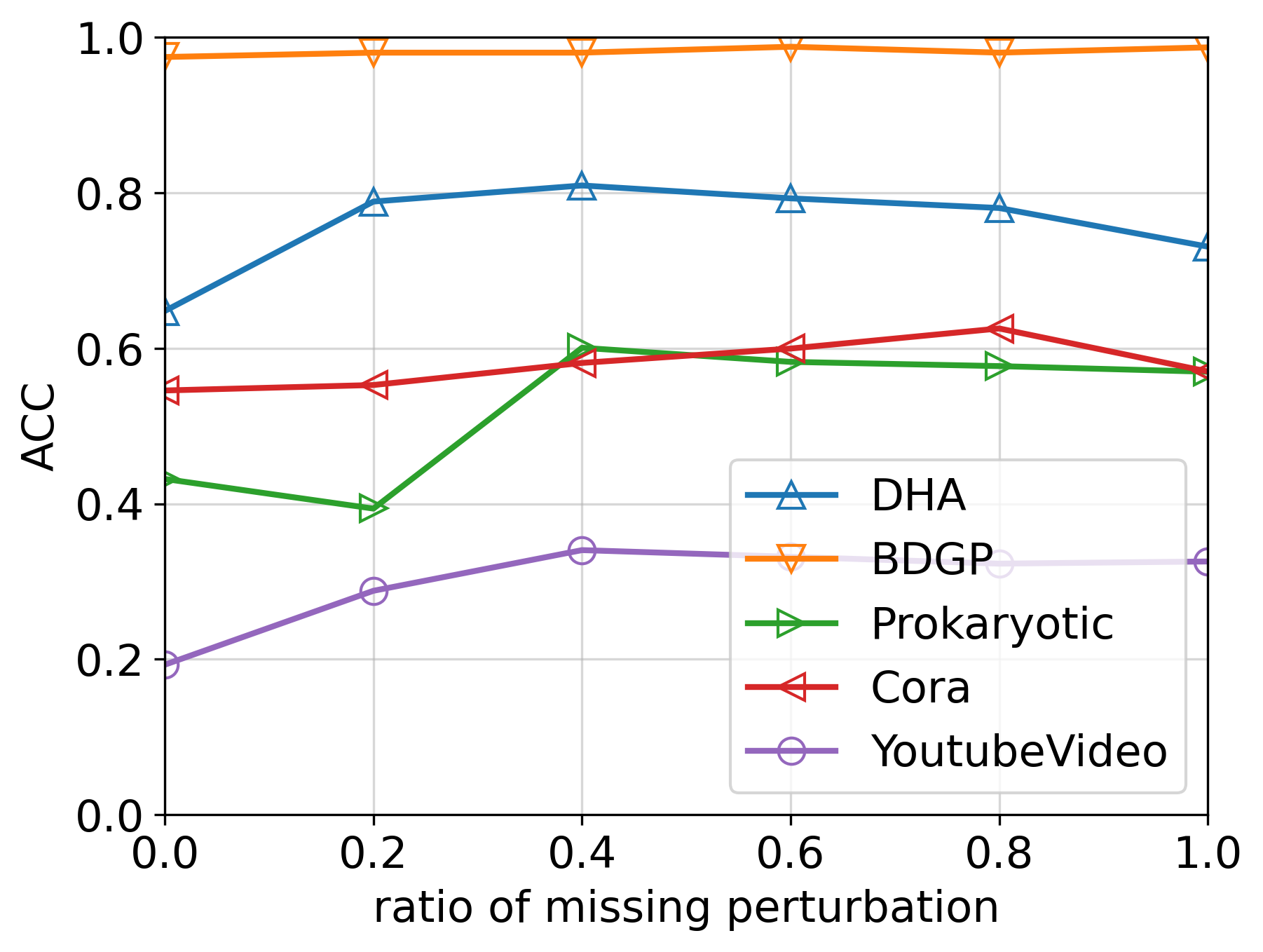}
\caption{$r$ $vs.$ ACC}
\vspace{-0.2cm}
\end{subfigure}
\caption{Hyper-parameter analysis of the different ratios on (a) noise perturbation and (b) unusable perturbation in RML.}\label{PRp}
\vspace{-0.3cm}
\end{figure}

In this section, we first present the ablation experiments of RML on different tasks.
Then, we conduct hyper-parameter analysis and visualization to understand its behavior.

Figure~\ref{abcp}, Table~\ref{table5p}, and Figure~\ref{abrp} present the ablation study on unsupervised multi-view clustering, noise-label multi-view classification, and cross-modal retrieval tasks, respectively.
Concretely, we let $\rm atten.$, $\rm NP$, and $\rm MP$ denote the sample-level attention, noise perturbation, and unusable perturbation, respectively.
$\rm base$ denotes the model without these three components and $\rm full$ is the complete RML model.
The results indicate that our proposed $\rm atten.$, $\rm NP$, and $\rm MP$ all contributed to improving the $\rm base$ model.
This demonstrates that the components in RML are effective and they have the potential to enhance different downstream tasks.

To observe the crucial hyper-parameters in our proposed simulated perturbations, we select the unsupervised multi-view clustering task and test $p$ and $r$ in the range of $[0.0, 1.0]$.
The results are depicted in Figure~\ref{PRp}, where we change one hyper-parameter with unchanged another.
For the two simulated perturbations, the experimental results suggest that the ratios being greater than $0.0$ and less than $1.0$ can enhance the model performance.
This observation is reasonable because setting $p$ and $r$ to $0.0$ would mean not using the simulated perturbations, while setting them to $1.0$ would result in excessive perturbations leading to huge information loss among multi-view data.
In comparison experiments, both $p$ and $r$ are set to $0.25$ on datasets DHA, BDGP, Cora, VOC, YoutubeVideo, Cifar100, MIRFLICKR-25K, NUS-WIDE; $0.50$ is on NGs and Prokaryotic; $0.75$ is on WebKB.

\begin{figure}[!t]
\centering
\begin{subfigure}{0.49\linewidth}
\includegraphics[width=\linewidth]{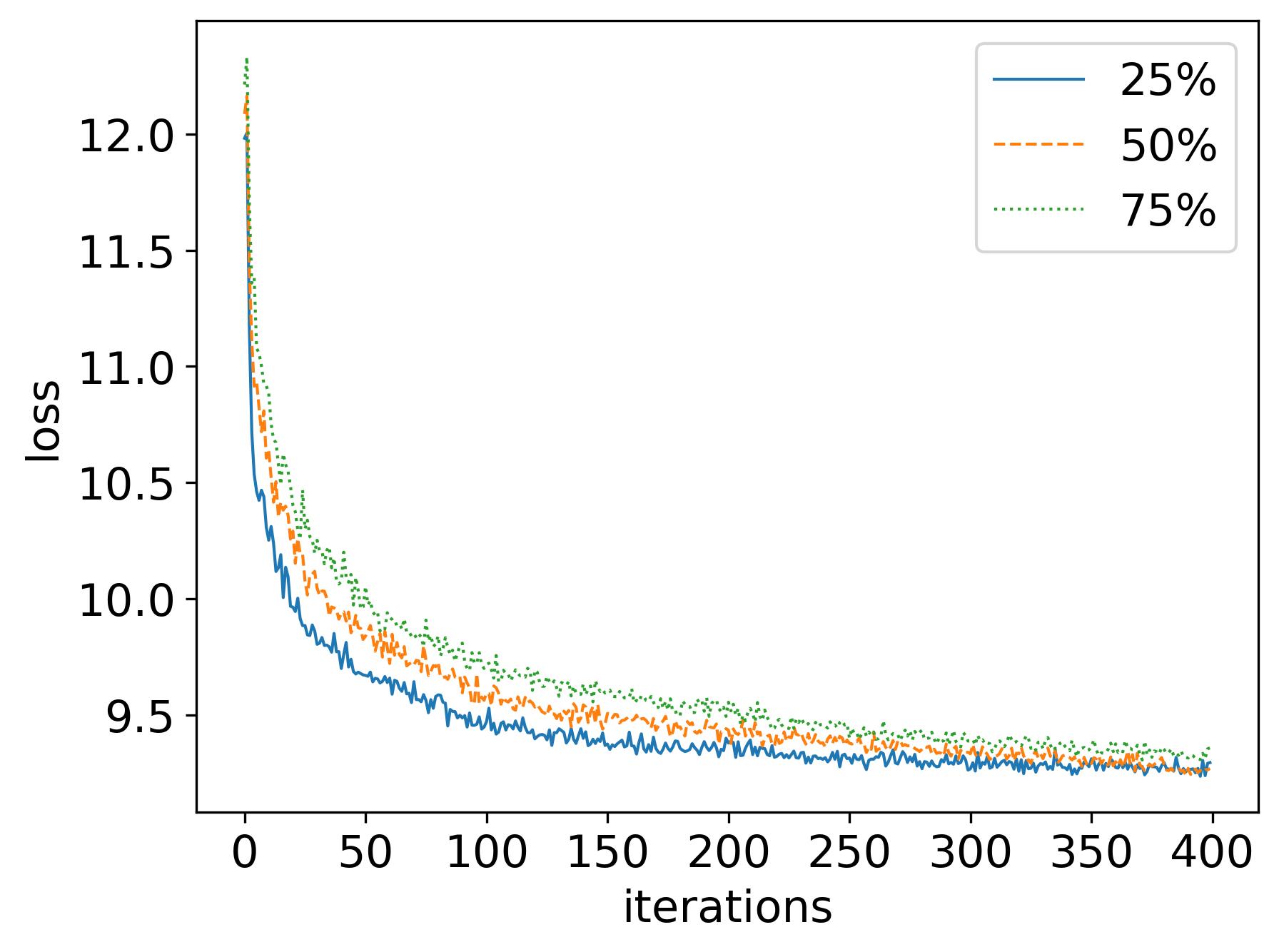}
\caption{DHA}
\vspace{-0.2cm}
\end{subfigure}
\begin{subfigure}{0.49\linewidth}
\includegraphics[width=\linewidth]{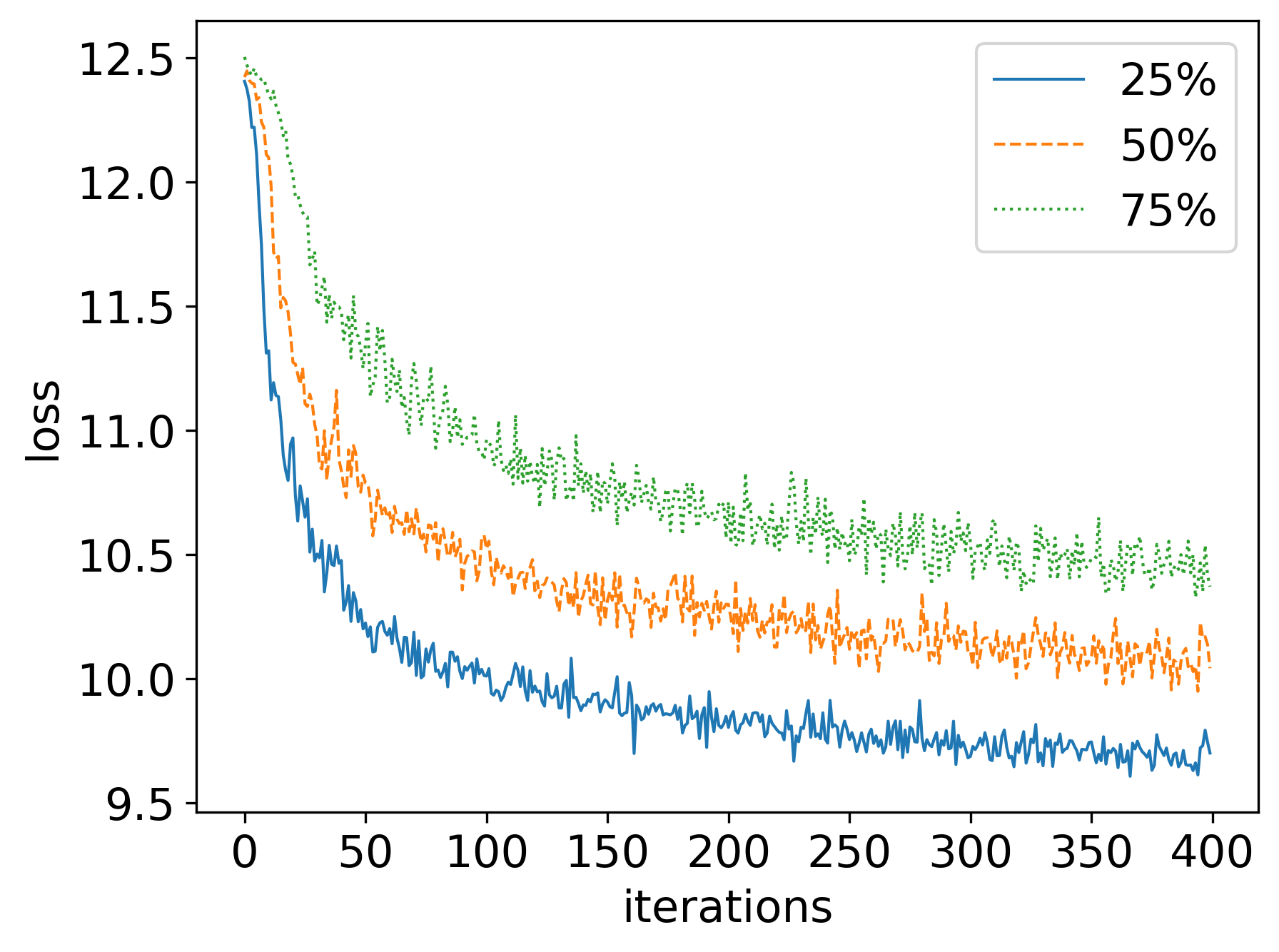}
\caption{BDGP}
\vspace{-0.2cm}
\end{subfigure}
\caption{The training loss values during our proposed simulated perturbation based multi-view contrastive learning.}\label{lossp}
\vspace{-0.3cm}
\end{figure}

To investigate the convergence of our proposed RML framework, we set the ratios of simulated perturbations to $25\%$, $50\%$, and $75\%$ (\ie, $p=r=0.25, 0.50, 0.75$), respectively, and visualize the variations of loss $\mathcal{L}_{\rm RML}$ during the representation learning process of RML as shown in Figure~\ref{lossp}.
Despite higher perturbation ratios increasing the loss values, our proposed simulated perturbation based multi-view contrastive learning together with the multi-view transformer fusion model exhibits good convergence across different multi-view datasets over different perturbation ratios.

\section{Conclusion}\label{CO}

In the literature, multi-view learning methods have achieved promising progress in fields such as image-text interactions.
However, the existing successful experiences are challenging to replicate in the data from like some medical or internet applications, due to the heterogeneous and imperfect nature of multi-view datasets in these areas.
In this paper, we propose a novel robust multi-view learning method RML, which is capable of learning fused representations to extract discriminative information from diverse multi-view datasets.
Our extensive experiments demonstrate that RML shows promising versatility and it can i) achieve effective multi-view fusion to enhance the unsupervised multi-view clustering, ii) increase the model robustness in noise-label multi-view classification, and iii) serve as a regularization term to facilitate cross-modal hashing retrieval tasks.
The future work is to extend the framework to more multi-view learning tasks.

\section*{Acknowledgment}
This research was supported in part by the National Key Research \& Development Program of China under Grant 2022YFA1004100, and in part by the Ministry of Education, Singapore, under its MOE Academic Research Fund Tier 2 (MOE-T2EP20124-0013).

{
    \small
    \bibliographystyle{ieeenat_fullname}
    \bibliography{JieXU}
}

\onecolumn

\section*{Appendix for Robust Multi-View Learning via Representation Fusion of Sample-Level Attention and Alignment of Simulated Perturbation}

~

\section{Related Work}\label{RW}

In this section, we discuss the connections and differences between our method and related work including multi-view learning, contrastive learning, and attention mechanism.

\subsection{Multi-view learning}

Multi-view learning (MVL) refers to models learning comprehensive information from multiple views with matched correspondences.
In this paper, we focus on deep learning based MVL methods and categorize existing methods into two types, \ie, representation fusion and representation alignment.
Representation fusion methods are the earliest popular in deep MVL, which aims to obtain a fused representation that is superior to representations of individual views \cite{abavisani2018deep,nagrani2021attention}.
Many of these methods produce more accurate results on the fused representation than that on individual views' representations, and use it to further refine their representation learning \cite{zhou2020end,xu2023self}.
Representation alignment methods are first investigated by canonical correlation analysis based deep MVL approaches \cite{andrew2013deep,yu2018category,sun2020learning}.
With the advancement of contrastive learning from self-supervised learning, an increasing number of deep MVL methods have adopted contrastive learning to capture the agreement across views \cite{tian2020contrastive,liu2021contrastive,10061269,trostenMVC,xu2022multi}.
To achieve the representation alignment, these contrastive MVL methods treat different views of a sample as positive pairs and maximize the similarity among their representations, thereby aiming to learn the semantic information across multiple views \cite{tang2022deepi,yan2023gcfagg,jin2023deep,chen2023deep}.
Different from previous deep MVL methods, our RML performs the sample-level attention based multi-view representation fusion, and then achieves the simulated perturbation based representation alignment between the fused representations rather than between views.

\subsection{Contrastive learning}

Contrastive learning is a validated and effective paradigm for self-supervised representation learning \cite{gutmann2010noise,schroff2015facenet}.
It usually constructs positive and negative sample pairs and encourages the model to learn discriminative representations, thereby aggregating the representations of positive sample pairs closer \cite{oord2018representation,chen2020simple}.
The approaches for constructing positive sample pairs vary according to the types of data.
For instance, in terms of image data, data augmentation techniques such as rotation and color filtering are typically employed to generate multiple images that are semantically consistent \cite{he2020momentum,grill2020bootstrap}.
For time-series data, adjacent samples in the sequence are used to construct positive sample pairs \cite{pan2021videomoco,qian2021spatiotemporal}.
Recently, contrastive learning has made significant progress in multi-view or multimodal domains, where different views or modalities of a sample are treated as positive sample pairs without the need for data augmentation \cite{tian2020contrastive,hassani2020contrastive,radford2021learning}.
Motivated by the success of data augmentation in contrastive learning~\cite{van2001art,wang2022contrastive,Luo2024SCM},
in this work, we propose a novel simulated perturbation based multi-view contrastive learning method for representation learning and downstream tasks, where the positive sample pairs are constructed by the two perturbed versions of fused representations.

\subsection{Attention mechanism}

Attention mechanism is an important technique initially introduced in the context of neural machine translation which enables models to selectively focus on relevant parts of the input data \cite{bahdanau2015neural,parikh2016decomposable}.
It computes a weighted sum of input features, where the weights are dynamically determined based on the relevance of each feature to the task at hand, and this allows models to handle dependencies more effectively than traditional methods.
Due to this property, attention mechanism has been integrated in many MVL applications \cite{nam2017dual,hori2017attention,zhou2020end}.
Transformer \cite{vaswani2017attention} is one of the most popular networks in deep learning, which is built upon the attention mechanism and excels at modeling long-range dependencies between elements in sequences.
Recent advances have also employed transformer-like networks to MVL \cite{sun2019videobert,radford2021learning,10123038}, where the goal usually is to integrate and process information from multiple views such as text, image, audio, and video.
However, the heterogeneous and imperfect natures of real-world multi-view data often hinder the transferability of existing successful experiences.
To this end, this work proposes a robust MVL method which has a sample-level attention based multi-view fusion model using a transformer-like encoder network.

\section{Implementation Details}
\subsection{Method details}

\quad ~For unsupervised multi-view clustering task, we directly utilize the model $\mathcal{F}_{\theta^f}$ and minimize the loss function $\mathcal{L}_{\rm RML}$. Then, we employ the unsupervised clustering algorithm K-means~\cite{hartigan1979algorithm} on the fused representations $\mathbf{Z}$ to obtain the clustering results.

For noise-label multi-view classification task, we extend our RML model $\mathcal{F}_{\theta^f}$ by adding a classification head $\mathcal{H}_{\omega}$, obtaining class prediction probabilities $\mathbf{q}_i = \mathcal{H}_{\omega}(\mathcal{F}_{\theta^f}(\{\mathbf{x}_i^m\}_{m=1}^V))$ through Softmax. Subsequently, we minimize the sum of $\mathcal{L}_{\rm RML}$ and cross-entropy loss on the training set. In this paper, we propose two variants for noise-label multi-view classification. The first one is formulated as follows:
\begin{equation}\label{C1task}
\begin{aligned}
&\mathcal{L} = \mathcal{L}_{\rm CE} + \lambda \mathcal{L}_{\rm RML} \\
&s.t.~\mathcal{L}_{\rm CE} = \mathcal{L}_{\rm CrossEntropy}(\mathbf{Y}, \{\mathbf{X}^m\}_{m=1}^V) \\
&~~~~~~~~~~~~~= - \sum_i \mathbf{y}_i \log \mathbf{q}_i.
\end{aligned}
\end{equation}
This variant is entitled as $\rm RML+\mathcal{L}_{\rm CE}$. Furthermore, we incorporate the proposed simulated perturbations to establish multiple cross-entropy objectives, for further improving the model robustness to imperfect multi-view data. To be specific, the second variant is defined as $\rm RML+\mathcal{L}_{\rm MCE}$:
\begin{equation}\label{C2task}
\begin{aligned}
&\mathcal{L} = \mathcal{L}_{\rm MCE} + \lambda \mathcal{L}_{\rm RML} \\
&s.t.~\mathcal{L}_{\rm MCE} = \mathcal{L}_{\rm CrossEntropy}(\mathbf{Y}, \{\mathbf{X}^m\}_{m=1}^V) \\
&~~~~~~~~~~~~~~~~+ \mathcal{L}_{\rm CrossEntropy}(\mathbf{Y}, \{\mathbf{N}^m\}_{m=1}^V) \\
&~~~~~~~~~~~~~~~~+ \mathcal{L}_{\rm CrossEntropy}(\mathbf{Y}, \{\mathbf{M}^m\}_{m=1}^V)  \\ 
&~~~~~~~~~~~~= - \sum_i \left (\mathbf{y}_i \log \mathbf{q}_i + \mathbf{y}_i \log \mathbf{q}_i^N + \mathbf{y}_i \log \mathbf{q}_i^M \right),
\end{aligned}
\end{equation}
where we have $\mathbf{q}_i^N = \mathcal{H}_{\omega}(\mathcal{F}_{\theta^f}(\{\mathbf{n}_i^m\}_{m=1}^V))$ and $\mathbf{q}_i^M = \mathcal{H}_{\omega}(\mathcal{F}_{\theta^f}(\{\mathbf{m}_i^m\}_{m=1}^V))$, by which we make the classification model more robust to the noise perturbed data $\mathbf{n}_i^m$ as well as the unusable perturbed data $\mathbf{m}_i^m$.

For cross-modal hashing retrieval task, we apply our method RML in a plug-and-play manner to existing cross-modal hashing retrieval approaches. Specifically, we integrate our RML model $\mathcal{F}_{\theta^f}$ on the top of the representation learning module of methods UCCH~\cite{hu2022unsupervised} and NRCH~\cite{wang2024robust}, and incorporate our optimization objective $\mathcal{L}_{\rm RML}$ as a regularization term into that of the cross-modal retrieval objective (\ie, $\mathcal{L}_{\rm UCCH}$ and $\mathcal{L}_{\rm NRCH}$):
\begin{equation}\label{Rtask}
\begin{aligned}
\mathcal{L} & = \mathcal{L}_{\rm UCCH} + \lambda \mathcal{L}_{\rm RML}, \\
\mathcal{L} & = \mathcal{L}_{\rm NRCH} + \lambda \mathcal{L}_{\rm RML}.
\end{aligned}
\end{equation}

\subsection{Experiment details}
In this paper, we established the common model architecture of RML for the three different tasks, \ie, multi-view clustering, multi-view classification, and cross-modal retrieval. This helps demonstrate the universality of our RML framework and promotes the comparable evaluation.
Specifically, we leverage MLP networks and attention layer to implement the multi-view transformer fusion network $\mathcal{F}_{\theta^f}$ in RML.
Firstly, $V$ parallel MLP networks are leveraged to transfer the input data $\{\mathbf{X}^m\}_{m=1}^V$ into word embeddings $\{\mathbf{E}^m\}_{m=1}^V$.
For the $m$-th view, the MLP network can be illustrated as $\mathbf{X}^m \rightarrow Fc(D_m)-GELU-dropout(0.2) \rightarrow Fc(D_m)-dropout(0.2) \rightarrow \mathbf{E}^m$, where $Fc(D_m)$ denotes the fully-connected network with $D_m$ neurons ($D_m$ is the data dimensionality of the $m$-th view), GELU is the active function of Gaussian Error Linear Unit \cite{hendrycks2016gaussian}, and $\rm dropout(0.2)$ is the dropout operation \cite{srivastava2014dropout} with the rate of $0.2$. Upon $\{\mathbf{E}^m\}_{m=1}^V$, we adopt the typical transformer encoder network to obtain $V$ encoded embeddings $\{\mathbf{F}^m\}_{m=1}^V$. Here, we use only one transformer encoder block \cite{vaswani2017attention} and the number of heads for multi-head attention is set to $1$. Finally, we add multiple $\{\mathbf{F}^m\}_{m=1}^V$ and utilize a one-layer fully-connected MLP network to achieve the fused representation $\mathbf{Z}$. The dimensions of $\{\mathbf{E}^m,\mathbf{F}^m\}_{m=1}^V$ and $\mathbf{Z}$ are all set to $256$ (\ie, $d_e$ and $d$ are set to $256$). We employ InfoNCE~\cite{oord2018representation} contrastive loss to implement the optimization objective $\mathcal{L}_{\rm RML}$, where the temperature $\tau$ is set to $0.5$. To train the model parameters, the optimizer we choose is Adam~\cite{Diederik} with the learning rate of $0.0003$.
$\sigma$ in the Gaussian distribution $\mathcal{N}(0, \sigma^2)$ is set to $0.4$.

When using K-Means clustering in our experiments, different views are concatenated to form a single one.
For a fair comparison, the hyper-parameters of all comparison methods adopted the recommended settings given by the authors, and these comparison methods use the same input multi-view or multimodal data as that used in our RML.

In our cross-modal retrieval experiments, we follow the experimental settings and results in UCCH~\cite{hu2022unsupervised} to evaluate the performance of baselines and our RML.
Specifically,
we conduct two kinds of cross-modal retrieval task, \ie, $\rm Image \rightarrow Text$ and $\rm Text \rightarrow Image$.
Here, the ground-truth relevant samples refer to the cross-modal samples which have the same semantic category as the query sample.
To evaluate the cross-modal retrieval results, the retrieval protocols adopt the same way in~\cite{hu2022unsupervised} that we measure the accuracy scores of the Hamming ranking results by Mean Average Precision (MAP), which returns the mean value of average precision scores for each query sample.
In our experiments, we take MAP@ALL where all MAP scores are calculated on all retrieval results returned by tested methods.
For RML+UCCH and RML+NRCH, to facilitate a fair comparison, we took the source code of UCCH and NRCH and inserted our RML module into them without introducing unnecessary changes.
Since NRCH has different settings in data partitioning and pre-processing from UCCH, we treat NRCH and RML+NRCH as another set of comparison.

\subsection{Dataset details}

As we highly expect a MVL method which is compatible with various multi-view datasets, we conducted experiments on multiple types of multi-view or multimodal datasets to validate the effectiveness and universality of methods. We provide the detailed information of datasets as follow:
\begin{itemize}
    \item \textbf{DHA}~\cite{lin2012human} is a repository documenting the intricacies of human motion, which captures RGB and depth image sequences as two views for each sample. Spanning across 23 unique categories, this multimodal dataset serves as a resource for the in-depth research of human motion.

    \item \textbf{BDGP}~\cite{cai2012joint} comprises 2,500 samples of drosophila embryos which are categorized into 5 different classes. For each sample, two views of features have been extracted, including a 1,750-dimensional visual feature and a 79-dimensional textual feature.

    \item \textbf{Prokaryotic}~\cite{brbic2016landscape} is a bioinformatics dataset that collects 551 prokaryotic species with three views.
    The dataset provides 4 species, described by textual features in the bag-of-words format, proteome compositions encoded by the frequency of amino acids, and gene repertoires using presence/absence indicators for gene families.

    \item \textbf{Cora}~\cite{bisson2012co} consists of 2,708 scientific documents published over 7 topics, such as neural networks, reinforcement learning, and theory.
    Each document has a content-citation pair, that is 1,433-dimensional word content information and 2,708-dimensional citation information.

    \item \textbf{YoutubeVideo}~\cite{madani2013using} is a large-scale multi-view dataset with 101,499 samples from 31 classes, in which 512-dimensional cuboids histogram, 647-dimensional HOG, and 838-dimensional MISC vision features are leveraged to describe video data collected from the YouTube website.

    \item \textbf{WebKB}~\cite{sun2007kernelized} is a dataset about web page information collected from the computer science departments of various universities. It comprises 1,051 samples belonging to course or non-course pages, and each sample has a fulltext view and an inlink view in web pages.

    \item \textbf{VOC}~\cite{everingham2010pascal} consists of image-text pairs to form a two-modality dataset, with 5,649 instances across 20 categories. For each sample, the first modality is represented by 512-dimensional image GIST features, while the second modality is characterized by a word frequency count of 399-dimensional features.
    
    \item \textbf{NGs}~\cite{hussain2010improved} is a subset of the newsgroup dataset, consisting of 500 newsgroup documents and 5 categories. Each document has three views obtained through pre-processing methods, \ie, supervised mutual information, partitioning around medoids, and unsupervised mutual information.

    \item \textbf{Cifar100}~\cite{krizhevsky2009learning} is a popular image database with 50,000 samples from 100 subcategories. We follow~\cite{9982492} that extracts the image features through ResNet18, ResNet50, and DenseNet121 to construct three views, respectively.

    \item \textbf{MIRFLICKR-25K}~\cite{huiskes2008mir} and \textbf{NUS-WIDE}~\cite{rasiwasia2010new} are two image-text datasets widely-used for cross-modal retrieval tasks (including image-to-text retrieval and text-to-image retrieval). We follow the setting in \cite{hu2022unsupervised} to ensure a fair comparison as follows.
    For MIRFLICKR-25K, 18,015 image-text pairs are randomly selected as the retrieval set and the left 2,000 pairs are used as the query set, where each sample is with multiple labels from 24 semantic categories.
    The pretrained 19-layer VGGNet extracts the 4,096-dimensional image features and the bag-of-words (BoW) obtains 1,386-dimensional text features.
    For NUS-WIDE, 184,457 image-text pairs are randomly selected as the retrieval set and the remaining 2,100 pairs are the query set, belonging to 10 classes.
    Each pair is represented by the 4,096-dimensional VGGNet image features and 1,000-dimensional BoW text features.
\end{itemize}

\section{More Experimental Results}
In this appendix, we provide more experimental results to support our claims in this paper.

For noise-label multi-view classification task, Table~\ref{table2} shows the results on different noise rates which further indicate the effectiveness of our RML to improve the robustness against noise labels. We provide the mean values of five independent runs of comparison experiments as well as the corresponding standard deviation in the following Tables~\ref{tablea2}, \ref{tableaa2}, and \ref{tableaa22}.
The results indicate that the improvement achieved by our method is significant.

\begin{table*}[!ht]
\caption{Performance comparison on noise-label multi-view classification}\label{table2}
\scriptsize
\centering
\resizebox{\textwidth}{!}{
\begin{threeparttable}
    \begin{tabular}{lccccccccccccccc}
    \toprule
    \multirow{1}{*}{Method} 
    &\multicolumn{3}{c}{DHA} &\multicolumn{3}{c}{BDGP} &\multicolumn{3}{c}{Prokaryotic} &\multicolumn{3}{c}{Cora} &\multicolumn{3}{c}{YoutubeVideo} \\
    \cmidrule(r){2-4} \cmidrule(r){5-7} \cmidrule(r){8-10} \cmidrule(r){11-13} \cmidrule(r){14-16}
    & ACC & Pre. & F1 & ACC & Pre. & F1 & ACC & Pre. & F1 & ACC & Pre. & F1 & ACC & Pre. & F1 \\
    
    \hline
    \multicolumn{16}{c}{noise label rate is 0\%} \\
    \hline
    Trans.+$\mathcal{L}_{\rm CE}$      &0.789 &0.829 &0.792 &0.967 &0.968 &0.967 &0.836 &0.841 &0.837 &0.828 &0.828 &0.827 &0.473 &0.740 &0.387 \\
    \rowcolor{gray!10}
    Trans.+$\mathcal{L}_{\rm MCE}$     &0.788 &0.819 &0.788 &0.903 &0.905 &0.903 &0.842 &0.850 &0.844 &0.778 &0.780 &0.778 &0.648 &0.711 &0.602 \\
    RML+$\mathcal{L}_{\rm CE}$         &0.712 &0.815 &0.670 &0.959 &0.959 &0.959 &0.854 &0.860 &0.855 &0.772 &0.775 &0.767 &0.759 &0.761 &0.758 \\
    \rowcolor{gray!10}
    RML+$\mathcal{L}_{\rm MCE}$        &0.796 &0.836 &0.795 &0.957 &0.958 &0.957 &0.852 &0.856 &0.853 &0.822 &0.828 &0.821 &0.773 &0.774 &0.773 \\
    
    \hline
    \multicolumn{16}{c}{noise label rate is 10\%} \\
    \hline
    Trans.+$\mathcal{L}_{\rm CE}$      &0.724 &0.770 &0.723 &0.845 &0.847 &0.845 &0.766 &0.778 &0.770 &0.753 &0.754 &0.753 &0.471 &0.725 &0.387 \\
    \rowcolor{gray!10}
    Trans.+$\mathcal{L}_{\rm MCE}$     &0.723 &0.764 &0.719 &0.789 &0.793 &0.789 &0.769 &0.780 &0.772 &0.720 &0.724 &0.719 &0.440 &0.762 &0.339 \\
    RML+$\mathcal{L}_{\rm CE}$         &0.688 &0.805 &0.640 &0.950 &0.951 &0.950 &0.795 &0.816 &0.801 &0.764 &0.767 &0.756 &0.754 &0.754 &0.753 \\
    \rowcolor{gray!10}
    RML+$\mathcal{L}_{\rm MCE}$        &0.727 &0.798 &0.710 &0.867 &0.868 &0.867 &0.776 &0.796 &0.782 &0.792 &0.797 &0.788 &0.766 &0.767 &0.765 \\
    
    \hline
    \multicolumn{16}{c}{noise label rate is 30\%} \\
    \hline
    Trans.+$\mathcal{L}_{\rm CE}$      &0.626 &0.676 &0.619 &0.605 &0.605 &0.603 &0.636 &0.680 &0.648 &0.577 &0.592 &0.580 &0.268 &0.804 &0.113 \\
    \rowcolor{gray!10}
    Trans.+$\mathcal{L}_{\rm MCE}$     &0.618 &0.656 &0.609 &0.600 &0.605 &0.599 &0.617 &0.687 &0.636 &0.548 &0.564 &0.551 &0.475 &0.706 &0.406 \\
    RML+$\mathcal{L}_{\rm CE}$         &0.622 &0.773 &0.568 &0.938 &0.938 &0.938 &0.769 &0.807 &0.778 &0.665 &0.673 &0.658 &0.590 &0.640 &0.580 \\
    \rowcolor{gray!10}
    RML+$\mathcal{L}_{\rm MCE}$        &0.623 &0.773 &0.570 &0.938 &0.939 &0.938 &0.767 &0.807 &0.777 &0.668 &0.678 &0.663 &0.600 &0.645 &0.593 \\
    
    \hline
    \multicolumn{16}{c}{noise label rate is 50\%} \\
    \hline
    Trans.+$\mathcal{L}_{\rm CE}$      &0.457 &0.487 &0.448 &0.437 &0.441 &0.435 &0.473 &0.594 &0.505 &0.400 &0.432 &0.407 &0.266 &0.804 &0.112 \\
    \rowcolor{gray!10}
    Trans.+$\mathcal{L}_{\rm MCE}$     &0.470 &0.519 &0.467 &0.442 &0.446 &0.441 &0.472 &0.606 &0.505 &0.374 &0.413 &0.382 &0.267 &0.805 &0.112 \\
    RML+$\mathcal{L}_{\rm CE}$         &0.608 &0.736 &0.563 &0.933 &0.933 &0.933 &0.735 &0.783 &0.747 &0.664 &0.669 &0.648 &0.592 &0.634 &0.584 \\
    \rowcolor{gray!10}
    RML+$\mathcal{L}_{\rm MCE}$        &0.610 &0.737 &0.565 &0.936 &0.936 &0.936 &0.735 &0.783 &0.747 &0.665 &0.666 &0.651 &0.598 &0.639 &0.593 \\

    \hline
    \multicolumn{16}{c}{noise label rate is 70\%} \\
    \hline
    Trans.+$\mathcal{L}_{\rm CE}$      &0.273 &0.309 &0.259 &0.256 &0.259 &0.255 &0.301 &0.477 &0.340 &0.269 &0.324 &0.282 &0.261 &0.637 &0.172 \\
    \rowcolor{gray!10}
    Trans.+$\mathcal{L}_{\rm MCE}$     &0.254 &0.275 &0.242 &0.249 &0.252 &0.249 &0.296 &0.470 &0.336 &0.259 &0.305 &0.271 &0.259 &0.512 &0.205 \\
    RML+$\mathcal{L}_{\rm CE}$         &0.421 &0.649 &0.330 &0.886 &0.890 &0.885 &0.402 &0.547 &0.437 &0.600 &0.630 &0.591 &0.586 &0.623 &0.580 \\
    \rowcolor{gray!10}
    RML+$\mathcal{L}_{\rm MCE}$        &0.422 &0.650 &0.331 &0.883 &0.887 &0.881 &0.408 &0.551 &0.443 &0.603 &0.622 &0.595 &0.587 &0.626 &0.580 \\
    \bottomrule
    \end{tabular}
\end{threeparttable}}
\end{table*}

\begin{table*}[!ht]
\caption{Performance comparison of unsupervised multi-view clustering on multi-view datasets (mean $\pm$ std)}\label{tablea2}
\small
\centering
\resizebox{\textwidth}{!}{
\begin{threeparttable}
    \begin{tabular}{lcccccccccccccccccccccccc}
    \toprule
    \multirow{1}{*}{Method} 
    &\multicolumn{2}{c}{DHA} &\multicolumn{2}{c}{BDGP} &\multicolumn{2}{c}{Prokaryotic} &\multicolumn{2}{c}{Cora} &\multicolumn{2}{c}{YoutubeVideo} \\
    \cmidrule(r){2-3} \cmidrule(r){4-5} \cmidrule(r){6-7} \cmidrule(r){8-9} \cmidrule(r){10-11}
    & ACC & NMI & ACC & NMI & ACC & NMI & ACC & NMI & ACC & NMI \\
    \hline
    K-means         &0.656±0.029 &0.798±0.001 &0.443±0.029 &0.573±0.041 &0.562±0.022 &0.325±0.006 &0.363±0.041 &0.172±0.043 &0.199±0.002 &0.194±0.001  \\
    
    MCN             &0.758±0.021 &0.800±0.017 &0.957±0.026 &0.901±0.041 &0.528±0.025 &0.287±0.014 &0.386±0.017 &0.184±0.032 &0.183±0.002 &0.187±0.001  \\
    
    CPSPAN          &0.663±0.033 &0.775±0.010 &0.690±0.087 &0.636±0.077 &0.539±0.031 &0.229±0.023 &0.419±0.030 &0.190±0.007 &0.232±0.014 &0.221±0.013  \\
    CVCL            &0.662±0.063 &0.754±0.033 &0.907±0.078 &0.785±0.009 &0.526±0.049 &0.281±0.032 &0.483±0.007 &0.310±0.003 &0.273±0.005 &0.258±0.002  \\
    DSIMVC          &0.635±0.046 &0.778±0.043 &0.983±0.003 &0.944±0.007 &0.597±0.017 &0.318±0.014 &0.478±0.037 &0.353±0.038 &0.189±0.003 &0.188±0.001  \\
    DSMVC           &0.762±0.013 &0.836±0.008 &0.523±0.079 &0.396±0.010 &0.502±0.063 &0.258±0.040 &0.447±0.041 &0.308±0.026 &0.178±0.002 &0.180±0.001  \\
    MFLVC           &0.716±0.011 &0.812±0.004 &0.983±0.012 &0.951±0.005 &0.569±0.034 &0.316±0.023 &0.485±0.041 &0.351±0.024 &0.184±0.002 &0.186±0.002  \\
    SCM             &0.814±0.021 &0.840±0.041 &0.962±0.003 &0.885±0.027 &0.550±0.030 &0.278±0.020 &0.564±0.020 &0.378±0.008 &0.316±0.007 &0.313±0.003  \\
    SCM$_{RE}$      &0.804±0.001 &0.840±0.001 &0.971±0.004 &0.913±0.002 &0.582±0.037 &0.312±0.028 &0.574±0.008 &0.374±0.009 &0.317±0.001 &0.322±0.004  \\
    \hline
    RML+K-means     &0.822±0.012 &0.847±0.005 &0.981±0.004 &0.941±0.009 &0.605±0.013 &0.316±0.014 &0.570±0.029 &0.371±0.011 &0.331±0.004 &0.339±0.003  \\
    \bottomrule
    \end{tabular}
\end{threeparttable}
}
\end{table*}

\begin{table*}[!ht]
\caption{Performance comparison of unsupervised multi-view clustering on multi-view datasets (mean $\pm$ std)}\label{tableaa2}
\scriptsize
\centering
\resizebox{\textwidth}{!}{
\begin{threeparttable}
    \begin{tabular}{lcccccccccccccccccccccccc}
    \toprule
    \multirow{1}{*}{Method} 
    &\multicolumn{2}{c}{WebKB} &\multicolumn{2}{c}{VOC} &\multicolumn{2}{c}{NGs} &\multicolumn{2}{c}{Cifar100} \\
    \cmidrule(r){2-3} \cmidrule(r){4-5} \cmidrule(r){6-7} \cmidrule(r){8-9}
    & ACC & NMI & ACC & NMI & ACC & NMI & ACC & NMI \\
    \hline
    K-means         &0.617±0.008 &0.002±0.001 &0.487±0.008 &0.360±0.020 &0.206±0.002 &0.019±0.003 &0.975±0.006 &0.996±0.001 \\
    
    MCN             &0.636±0.002 &0.081±0.002 &0.274±0.035 &0.286±0.011 &0.886±0.006 &0.736±0.002 &0.864±0.023 &0.962±0.001 \\
    
    CPSPAN          &0.771±0.021 &0.166±0.042 &0.452±0.022 &0.488±0.017 &0.352±0.002 &0.215±0.015 &0.918±0.014 &0.982±0.002 \\
    CVCL            &0.741±0.030 &0.246±0.026 &0.315±0.041 &0.317±0.026 &0.568±0.077 &0.317±0.078 &0.956±0.003 &0.977±0.001 \\
    DSIMVC          &0.702±0.014 &0.250±0.013 &0.212±0.017 &0.204±0.011 &0.630±0.062 &0.502±0.059 &0.895±0.011 &0.969±0.005 \\
    DSMVC           &0.663±0.018 &0.134±0.012 &0.633±0.034 &0.723±0.041 &0.352±0.027 &0.082±0.013 &0.851±0.023 &0.959±0.007 \\
    MFLVC           &0.672±0.021 &0.245±0.014 &0.292±0.004 &0.280±0.001 &0.908±0.000 &0.802±0.000 &0.877±0.018 &0.964±0.009 \\
    SCM             &0.689±0.017 &0.094±0.021 &0.607±0.046 &0.622±0.043 &0.968±0.004 &0.900±0.012 &0.999±0.001 &0.999±0.000 \\
    SCM$_{RE}$      &0.725±0.024 &0.268±0.052 &0.629±0.001 &0.629±0.011 &0.965±0.001 &0.893±0.001 &0.999±0.000 &0.999±0.000 \\
    \hline
    RML+K-means     &0.868±0.079 &0.508±0.156 &0.656±0.031 &0.615±0.011 &0.983±0.007 &0.943±0.022 &0.999±0.000 &0.999±0.000 \\
    \bottomrule
    \end{tabular}
\end{threeparttable}
}
\end{table*}

\begin{table*}[!ht]
\caption{Performance comparison on noise-label multi-view classification (mean $\pm$ std)}\label{tableaa22}
\small
\centering
\resizebox{\textwidth}{!}{
\begin{threeparttable}
    \begin{tabular}{lccccccccccccccc}
    \toprule
    \multirow{1}{*}{Method} 
    &\multicolumn{3}{c}{DHA} &\multicolumn{3}{c}{BDGP} &\multicolumn{3}{c}{Prokaryotic} &\multicolumn{3}{c}{Cora} &\multicolumn{3}{c}{YoutubeVideo} \\
    \cmidrule(r){2-4} \cmidrule(r){5-7} \cmidrule(r){8-10} \cmidrule(r){11-13} \cmidrule(r){14-16}
    & ACC & Pre. & F1 & ACC & Pre. & F1 & ACC & Pre. & F1 & ACC & Pre. & F1 & ACC & Pre. & F1 \\
    
    \hline
    \multicolumn{16}{c}{noise label rate is 0\%} \\
    \hline
    Trans.+$\mathcal{L}_{\rm CE}$      &0.789±0.023 &0.829±0.025 &0.792±0.025 &0.967±0.013 &0.968±0.013 &0.967±0.014 &0.836±0.013 &0.841±0.008 &0.837±0.011 &0.828±0.006 &0.828±0.005 &0.827±0.006 &0.473±0.170 &0.740±0.054 &0.387±0.225 \\
    Trans.+$\mathcal{L}_{\rm MCE}$     &0.788±0.034 &0.819±0.037 &0.788±0.035 &0.903±0.010 &0.905±0.010 &0.903±0.010 &0.842±0.019 &0.850±0.014 &0.844±0.017 &0.778±0.009 &0.780±0.009 &0.778±0.009 &0.648±0.020 &0.711±0.007 &0.602±0.028 \\
    RML+$\mathcal{L}_{\rm CE}$  &0.712±0.036 &0.815±0.031 &0.670±0.047 &0.959±0.006 &0.959±0.006 &0.959±0.006 &0.854±0.025 &0.860±0.018 &0.855±0.023 &0.772±0.012 &0.775±0.012 &0.767±0.015 &0.759±0.003 &0.761±0.003 &0.758±0.003 \\
    RML+$\mathcal{L}_{\rm MCE}$ &0.796±0.027 &0.836±0.020 &0.795±0.028 &0.957±0.007 &0.958±0.007 &0.957±0.007 &0.852±0.022 &0.856±0.016 &0.853±0.020 &0.822±0.016 &0.828±0.014 &0.821±0.017 &0.773±0.002 &0.774±0.003 &0.773±0.002 \\
    
    \hline
    \multicolumn{16}{c}{noise label rate is 10\%} \\
    \hline
    Trans.+$\mathcal{L}_{\rm CE}$      &0.724±0.036 &0.770±0.027 &0.723±0.042 &0.845±0.022 &0.847±0.021 &0.845±0.022 &0.766±0.023 &0.778±0.017 &0.770±0.021 &0.753±0.015 &0.754±0.016 &0.753±0.015 &0.471±0.167 &0.725±0.065 &0.387±0.224 \\
    Trans.+$\mathcal{L}_{\rm MCE}$     &0.723±0.027 &0.764±0.022 &0.719±0.033 &0.789±0.018 &0.793±0.018 &0.789±0.018 &0.769±0.021 &0.780±0.017 &0.772±0.019 &0.720±0.014 &0.724±0.015 &0.719±0.015 &0.440±0.201 &0.762±0.050 &0.339±0.263 \\
    RML+$\mathcal{L}_{\rm CE}$  &0.688±0.031 &0.805±0.036 &0.640±0.045 &0.950±0.013 &0.951±0.013 &0.950±0.013 &0.795±0.021 &0.816±0.010 &0.801±0.017 &0.764±0.006 &0.767±0.008 &0.756±0.013 &0.754±0.006 &0.754±0.006 &0.753±0.006 \\
    RML+$\mathcal{L}_{\rm MCE}$ &0.727±0.027 &0.798±0.037 &0.710±0.043 &0.867±0.023 &0.868±0.024 &0.867±0.024 &0.776±0.013 &0.796±0.004 &0.782±0.010 &0.792±0.021 &0.797±0.023 &0.788±0.027 &0.766±0.003 &0.767±0.003 &0.765±0.003 \\
    
    \hline
    \multicolumn{16}{c}{noise label rate is 30\%} \\
    \hline
    Trans.+$\mathcal{L}_{\rm CE}$      &0.626±0.073 &0.676±0.075 &0.619±0.074 &0.605±0.021 &0.605±0.016 &0.603±0.018 &0.636±0.045 &0.680±0.034 &0.648±0.041 &0.577±0.017 &0.592±0.013 &0.580±0.016 &0.268±0.001 &0.804±0.001 &0.113±0.001 \\
    Trans.+$\mathcal{L}_{\rm MCE}$     &0.618±0.044 &0.656±0.043 &0.609±0.044 &0.600±0.039 &0.605±0.038 &0.599±0.039 &0.617±0.047 &0.687±0.047 &0.636±0.045 &0.548±0.015 &0.564±0.016 &0.551±0.015 &0.475±0.171 &0.706±0.081 &0.406±0.241 \\
    RML+$\mathcal{L}_{\rm CE}$  &0.622±0.009 &0.773±0.023 &0.568±0.019 &0.938±0.008 &0.938±0.007 &0.938±0.008 &0.769±0.035 &0.807±0.022 &0.778±0.032 &0.665±0.015 &0.673±0.020 &0.658±0.017 &0.590±0.014 &0.640±0.003 &0.580±0.020 \\
    RML+$\mathcal{L}_{\rm MCE}$ &0.623±0.010 &0.773±0.022 &0.570±0.023 &0.938±0.006 &0.939±0.006 &0.938±0.006 &0.767±0.035 &0.807±0.022 &0.777±0.031 &0.668±0.014 &0.678±0.016 &0.663±0.015 &0.600±0.007 &0.645±0.007 &0.593±0.013 \\
    
    \hline
    \multicolumn{16}{c}{noise label rate is 50\%} \\
    \hline
    Trans.+$\mathcal{L}_{\rm CE}$      &0.457±0.065 &0.487±0.077 &0.448±0.073 &0.437±0.025 &0.441±0.027 &0.435±0.024 &0.473±0.052 &0.594±0.026 &0.505±0.043 &0.400±0.016 &0.432±0.016 &0.407±0.017 &0.266±0.001 &0.804±0.001 &0.112±0.001 \\
    Trans.+$\mathcal{L}_{\rm MCE}$     &0.470±0.043 &0.519±0.033 &0.467±0.043 &0.442±0.026 &0.446±0.028 &0.441±0.027 &0.472±0.048 &0.606±0.037 &0.505±0.040 &0.374±0.012 &0.413±0.015 &0.382±0.011 &0.267±0.002 &0.805±0.001 &0.112±0.001 \\
    RML+$\mathcal{L}_{\rm CE}$  &0.608±0.027 &0.736±0.022 &0.563±0.035 &0.933±0.014 &0.933±0.013 &0.933±0.014 &0.735±0.019 &0.783±0.020 &0.747±0.018 &0.664±0.011 &0.669±0.013 &0.648±0.012 &0.592±0.005 &0.634±0.005 &0.584±0.010 \\
    RML+$\mathcal{L}_{\rm MCE}$ &0.610±0.029 &0.737±0.025 &0.565±0.038 &0.936±0.009 &0.936±0.008 &0.936±0.009 &0.735±0.019 &0.783±0.020 &0.747±0.018 &0.665±0.014 &0.666±0.019 &0.651±0.013 &0.598±0.004 &0.639±0.007 &0.593±0.007 \\

    \hline
    \multicolumn{16}{c}{noise label rate is 70\%} \\
    \hline
    Trans.+$\mathcal{L}_{\rm CE}$      &0.273±0.049 &0.309±0.059 &0.259±0.050 &0.256±0.021 &0.259±0.022 &0.255±0.021 &0.301±0.035 &0.477±0.038 &0.340±0.032 &0.269±0.010 &0.324±0.016 &0.282±0.010 &0.261±0.006 &0.637±0.206 &0.172±0.074 \\
    Trans.+$\mathcal{L}_{\rm MCE}$     &0.254±0.060 &0.275±0.072 &0.242±0.061 &0.249±0.016 &0.252±0.019 &0.249±0.018 &0.296±0.018 &0.470±0.016 &0.336±0.013 &0.259±0.008 &0.305±0.014 &0.271±0.008 &0.259±0.007 &0.512±0.239 &0.205±0.076 \\
    RML+$\mathcal{L}_{\rm CE}$  &0.421±0.017 &0.649±0.030 &0.330±0.028 &0.886±0.041 &0.890±0.042 &0.885±0.044 &0.402±0.040 &0.547±0.051 &0.437±0.035 &0.600±0.017 &0.630±0.023 &0.591±0.014 &0.586±0.007 &0.623±0.004 &0.580±0.012 \\
    RML+$\mathcal{L}_{\rm MCE}$ &0.422±0.015 &0.650±0.030 &0.331±0.028 &0.883±0.051 &0.887±0.052 &0.881±0.054 &0.408±0.038 &0.551±0.050 &0.443±0.033 &0.603±0.011 &0.622±0.019 &0.595±0.014 &0.587±0.005 &0.626±0.007 &0.580±0.010 \\

    
    \bottomrule
    \end{tabular}
\end{threeparttable}}
\end{table*}

\begin{figure*}[!ht]
\centering
\begin{subfigure}{0.245\linewidth}
\includegraphics[width=\linewidth]{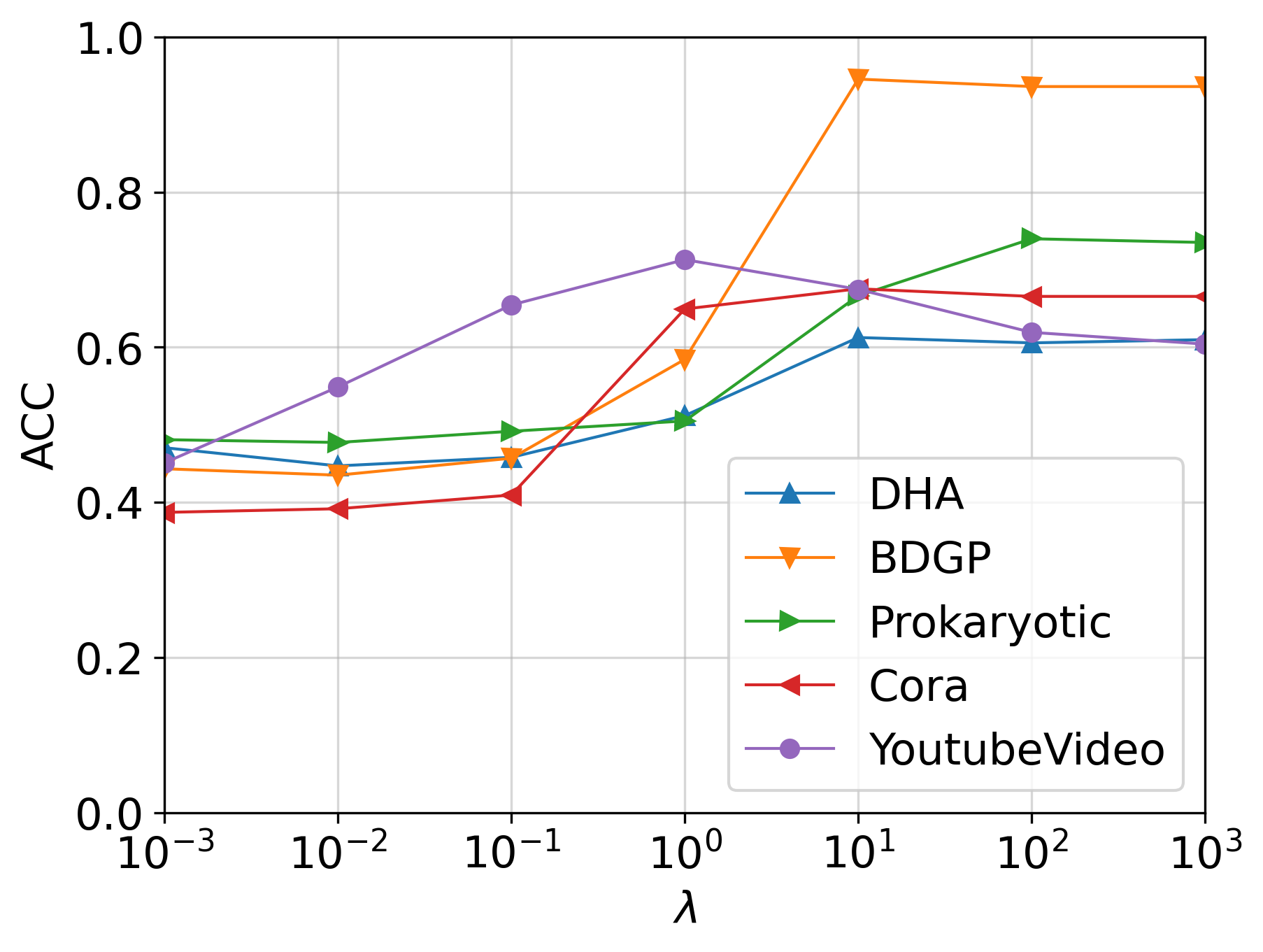}
\caption{ACC}
\end{subfigure} ~~~~~~
\begin{subfigure}{0.245\linewidth}
\includegraphics[width=\linewidth]{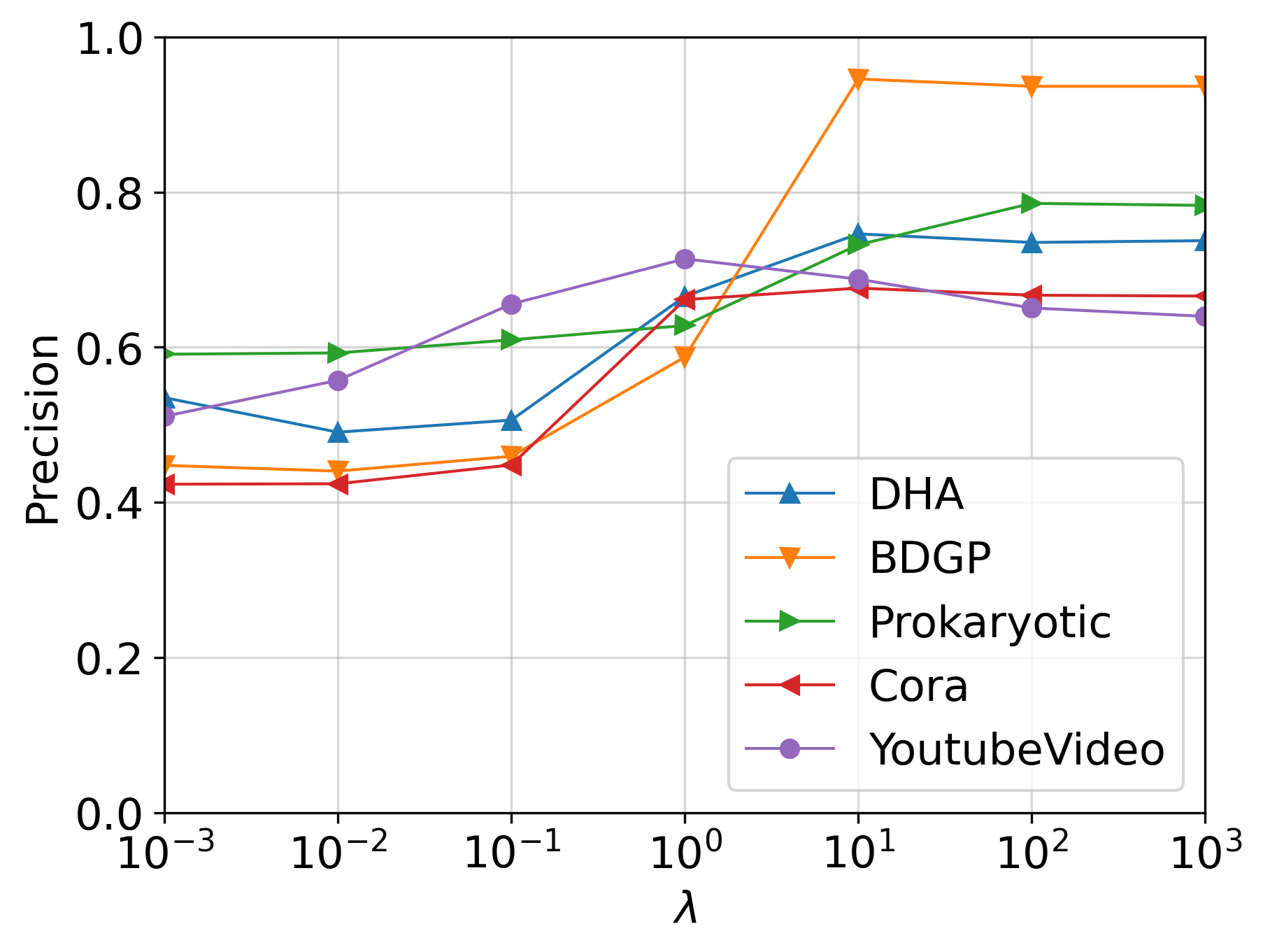}
\caption{Pre.}
\end{subfigure} ~~~~~~
\begin{subfigure}{0.245\linewidth}
\includegraphics[width=\linewidth]{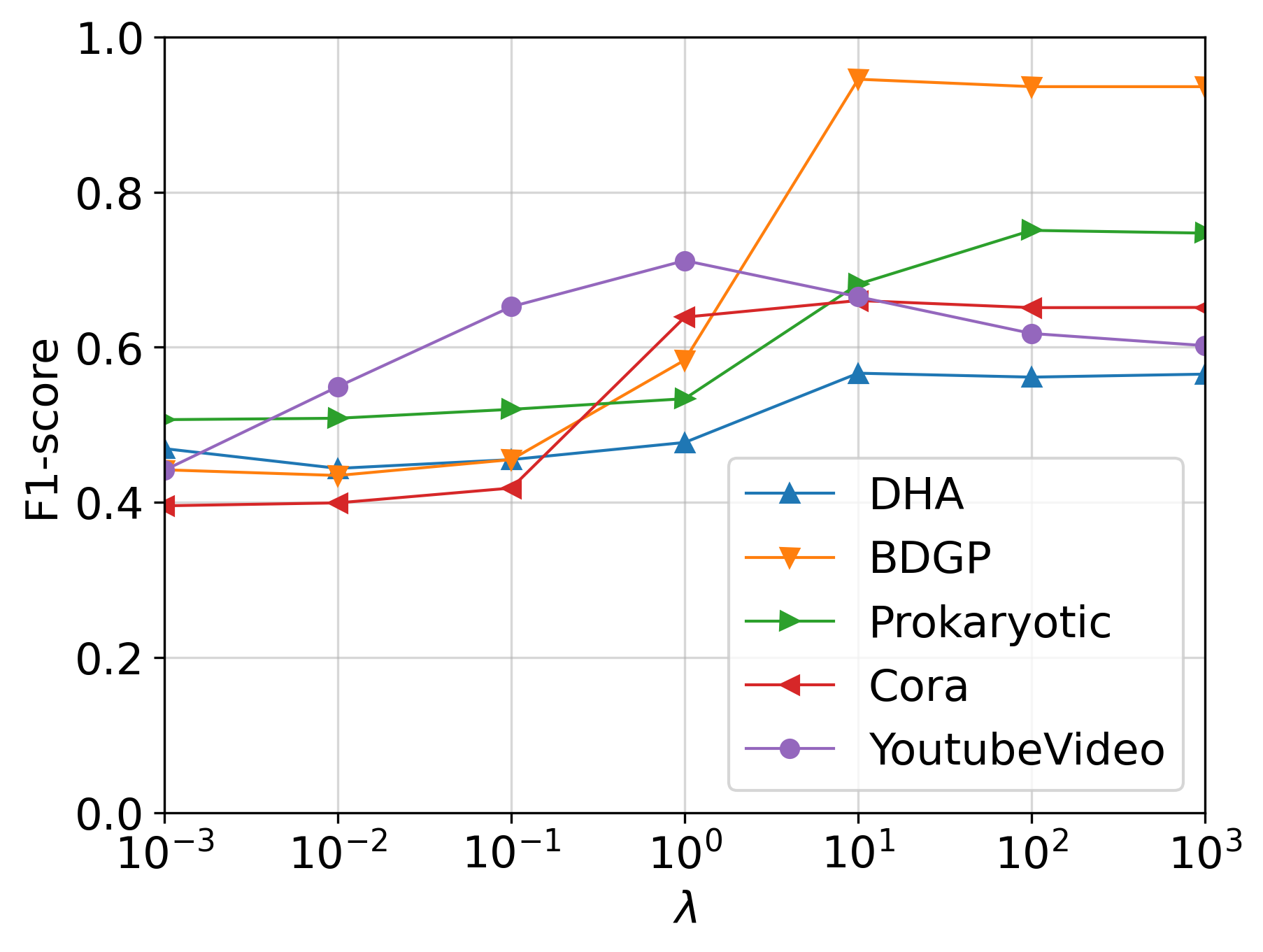}
\caption{F1}
\end{subfigure}
\caption{The hyper-parameter analysis of $\lambda$ over three metrics on noise-label multi-view classification with the noise label rate of $50\%$.}\label{LN}
\end{figure*}

\begin{figure*}[!ht]
\centering
\begin{subfigure}{.245\linewidth}
\includegraphics[width=\linewidth]{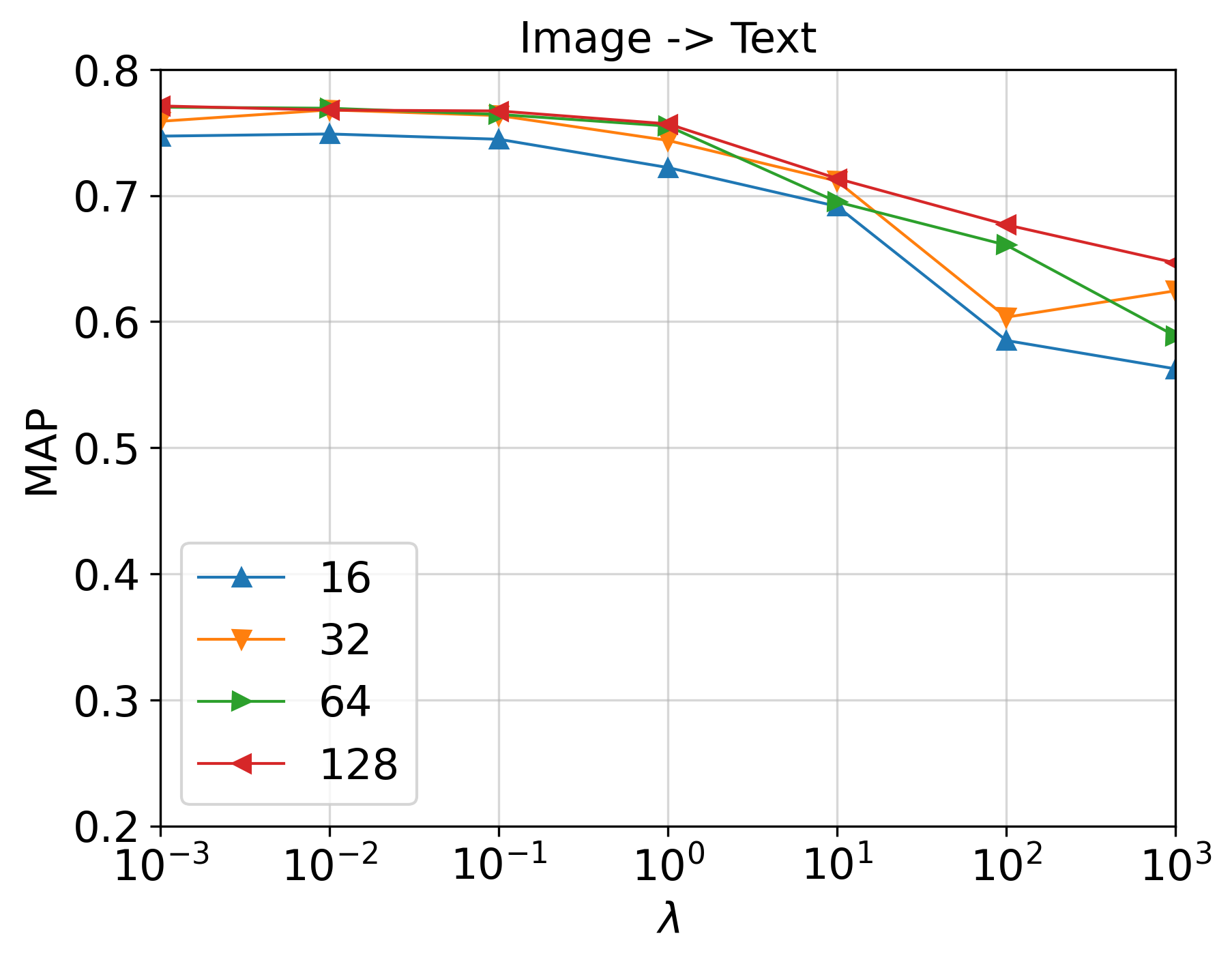}
\caption{MIRFLICKR-25K (image-to-text)}
\end{subfigure}
\begin{subfigure}{.245\linewidth}
\includegraphics[width=\linewidth]{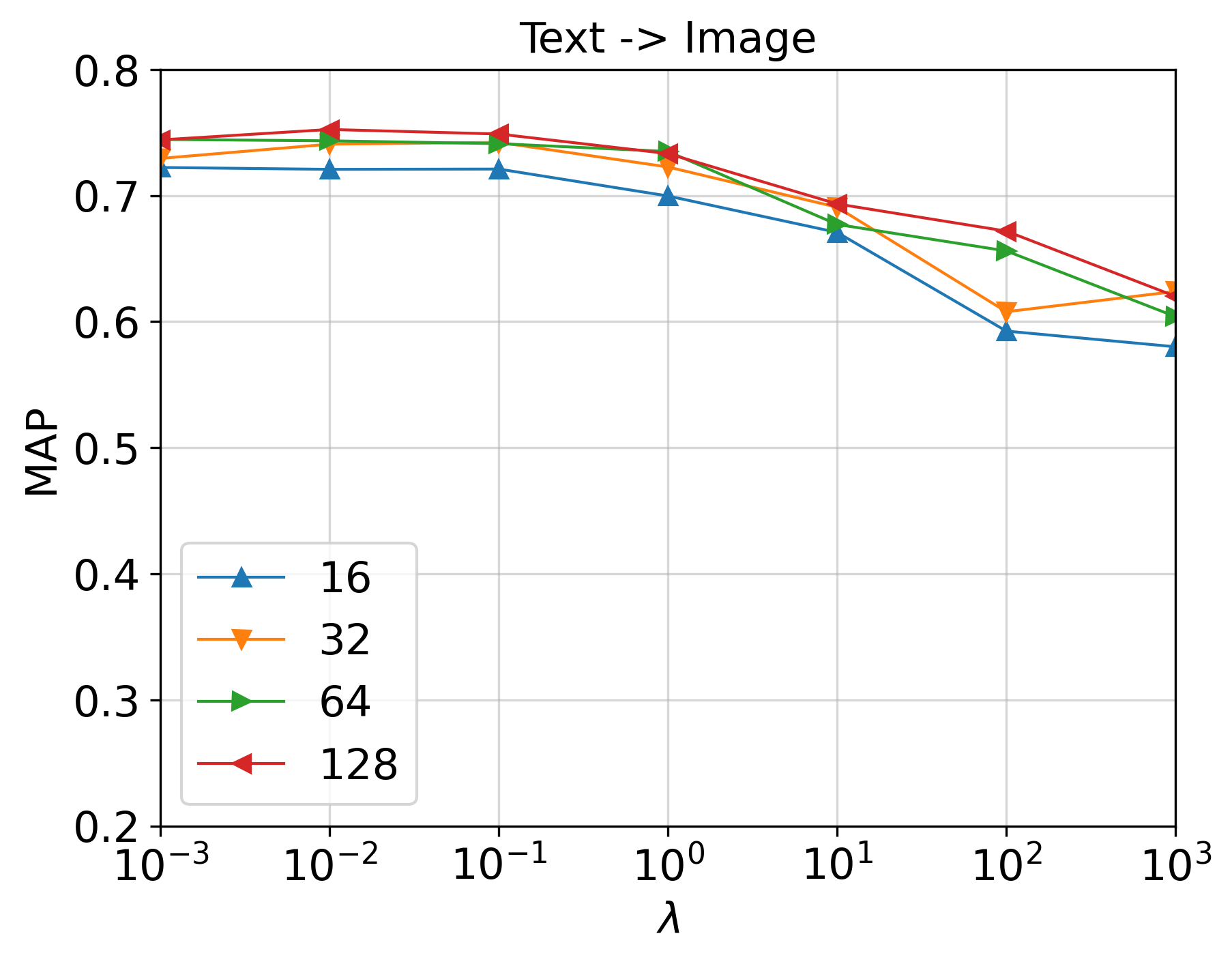}
\caption{MIRFLICKR-25K (text-to-image)}
\end{subfigure}
\begin{subfigure}{.245\linewidth}
\includegraphics[width=\linewidth]{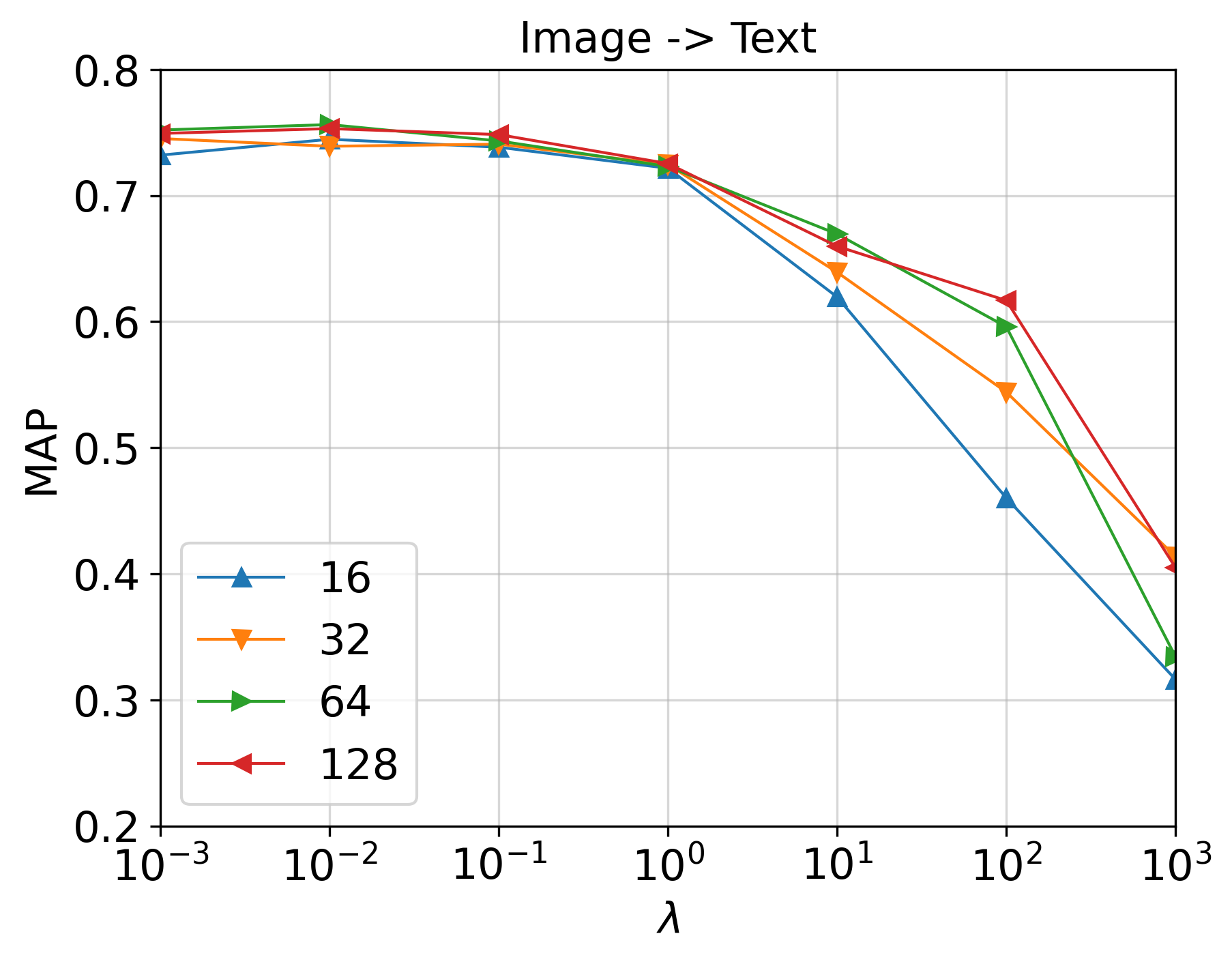}
\caption{NUS-WIDE (image-to-text)}
\end{subfigure}
\begin{subfigure}{.245\linewidth}
\includegraphics[width=\linewidth]{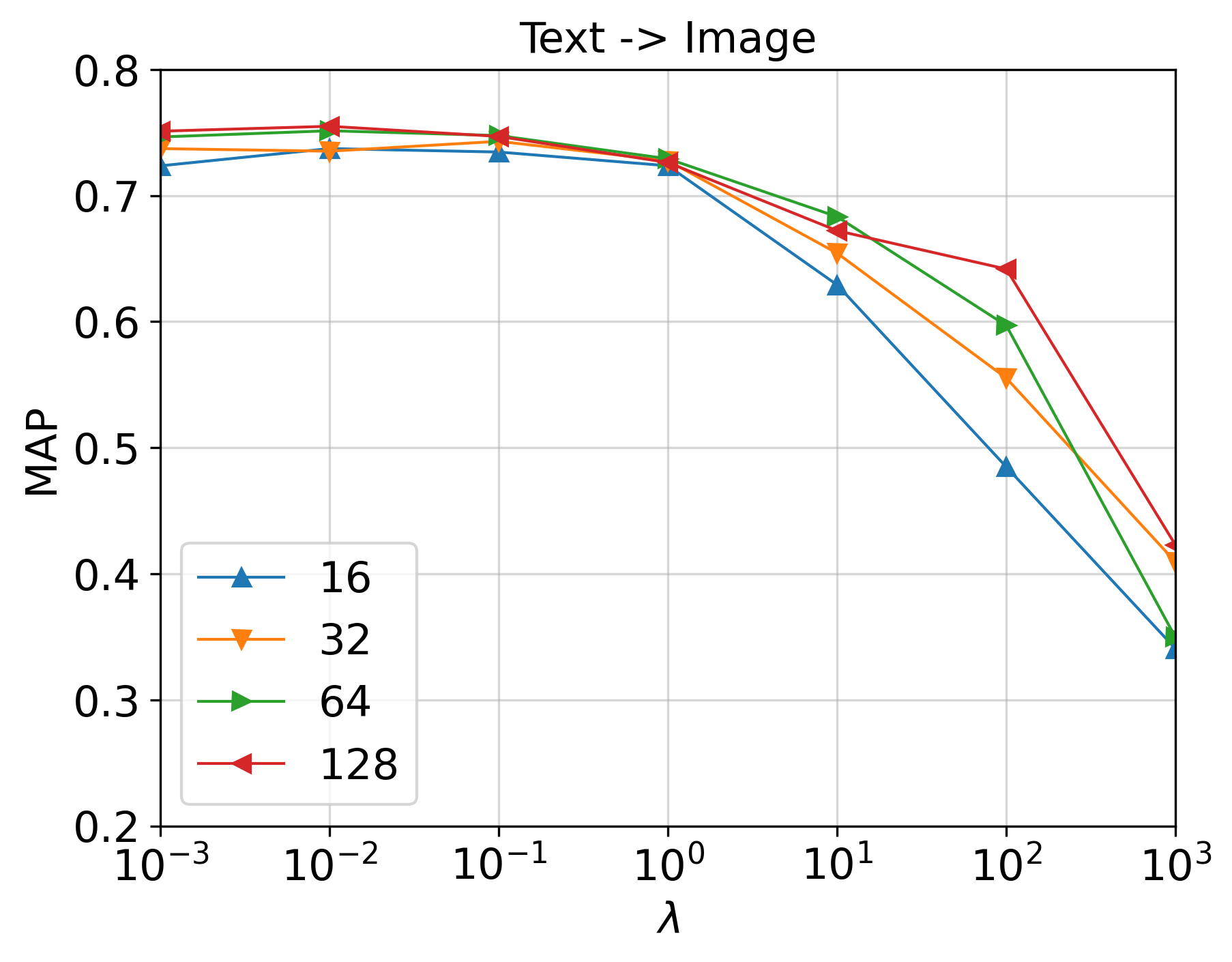}
\caption{NUS-WIDE (text-to-image)}
\end{subfigure}
\caption{The hyper-parameter analysis of $\lambda$ on cross-modal hashing retrieval tasks over hash code lengths of $[16,32,64,128]$, including image-to-text retrieval (a,c) and text-to-image retrieval (b,d) on datasets MIRFLICKR-25K and NUS-WIDE.}\label{LR}
\end{figure*}

Regarding hyper-parameter $\lambda$, we consider noise-label multi-view classification and cross-modal hashing retrieval tasks, where $\mathcal{L}_{\rm RML}$ is treated as a regularization term weighted by $\lambda$.
The parameter analysis with the noise label rate of $50\%$ is shown in Figure~\ref{LN}, where we observe stable classification performance within the range of $[10^1, 10^2, 10^3]$.
For the noise-label multi-view classification task, $\lambda$ is set to $10^3$ to emphasize $\mathcal{L}_{\rm RML}$ in joint optimization when the noise label rates are large (\eg, $30\%$, $50\%$, $70\%$).
When the noise label rates are small (\eg, $0\%$, $10\%$), $\lambda$ is set to $10^0$ for recommended settings.
For the cross-modal hashing retrieval tasks, stable performance is observed within the range of $[10^{-3}, 10^{-2}, 10^{-1}]$ as shown in Figure~\ref{LR}.
On cross-modal retrieval datasets MIRFLICKR-25K and NUS-WIDE, we kept $\lambda$ unchanged in our comparison experiments (\ie, $\lambda = 10^{-1}$).

\begin{figure*}[!ht]
\centering
\begin{subfigure}{0.245\linewidth}
\includegraphics[width=\linewidth]{Figure/NR_ACC.png}
\caption{$p$ $vs.$ ACC}
\end{subfigure}
\begin{subfigure}{0.245\linewidth}
\includegraphics[width=\linewidth]{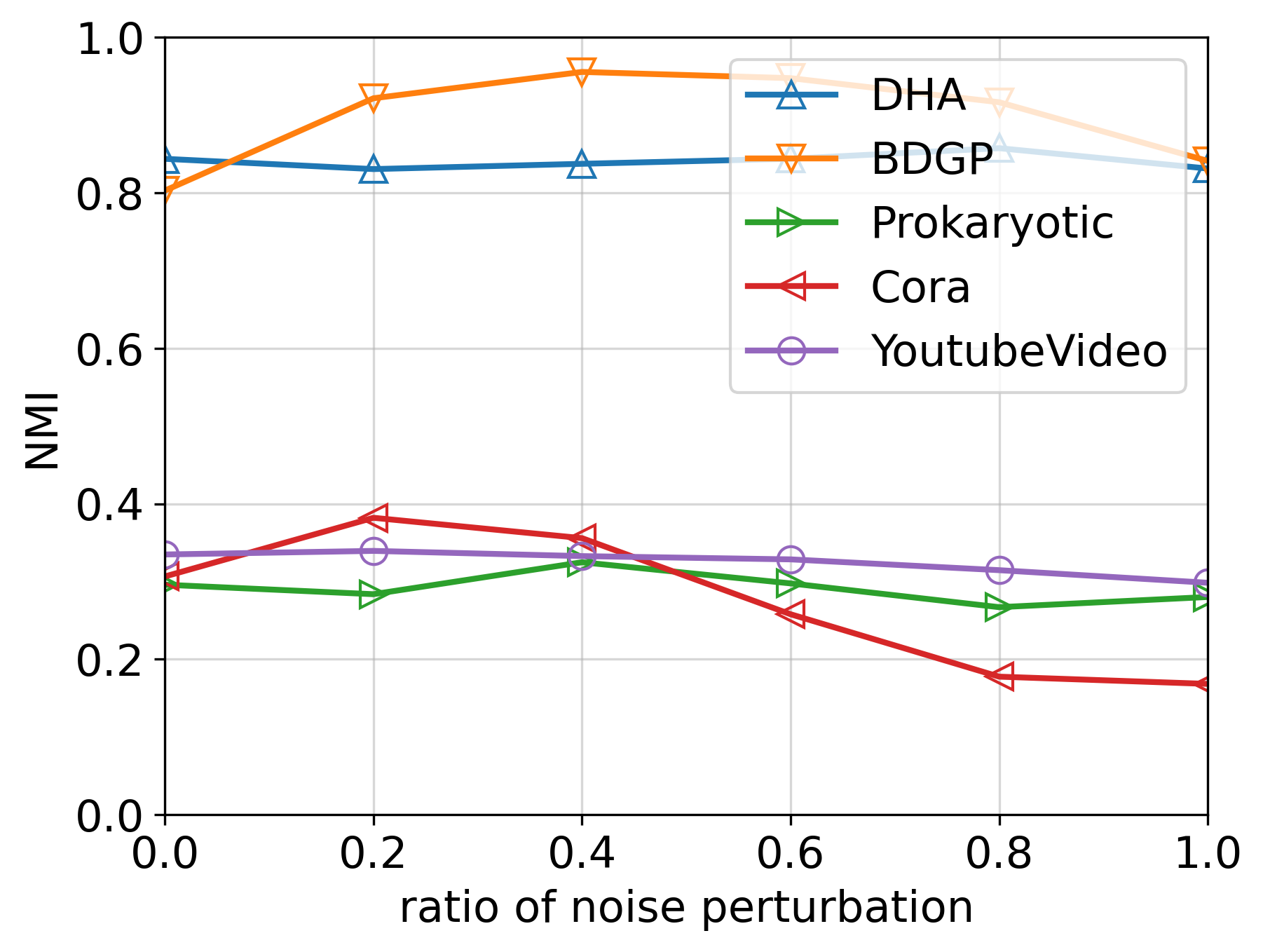}
\caption{$p$ $vs.$ NMI}
\end{subfigure}
\begin{subfigure}{0.245\linewidth}
\includegraphics[width=\linewidth]{Figure/MR_ACC.png}
\caption{$r$ $vs.$ ACC}
\end{subfigure}
\begin{subfigure}{0.245\linewidth}
\includegraphics[width=\linewidth]{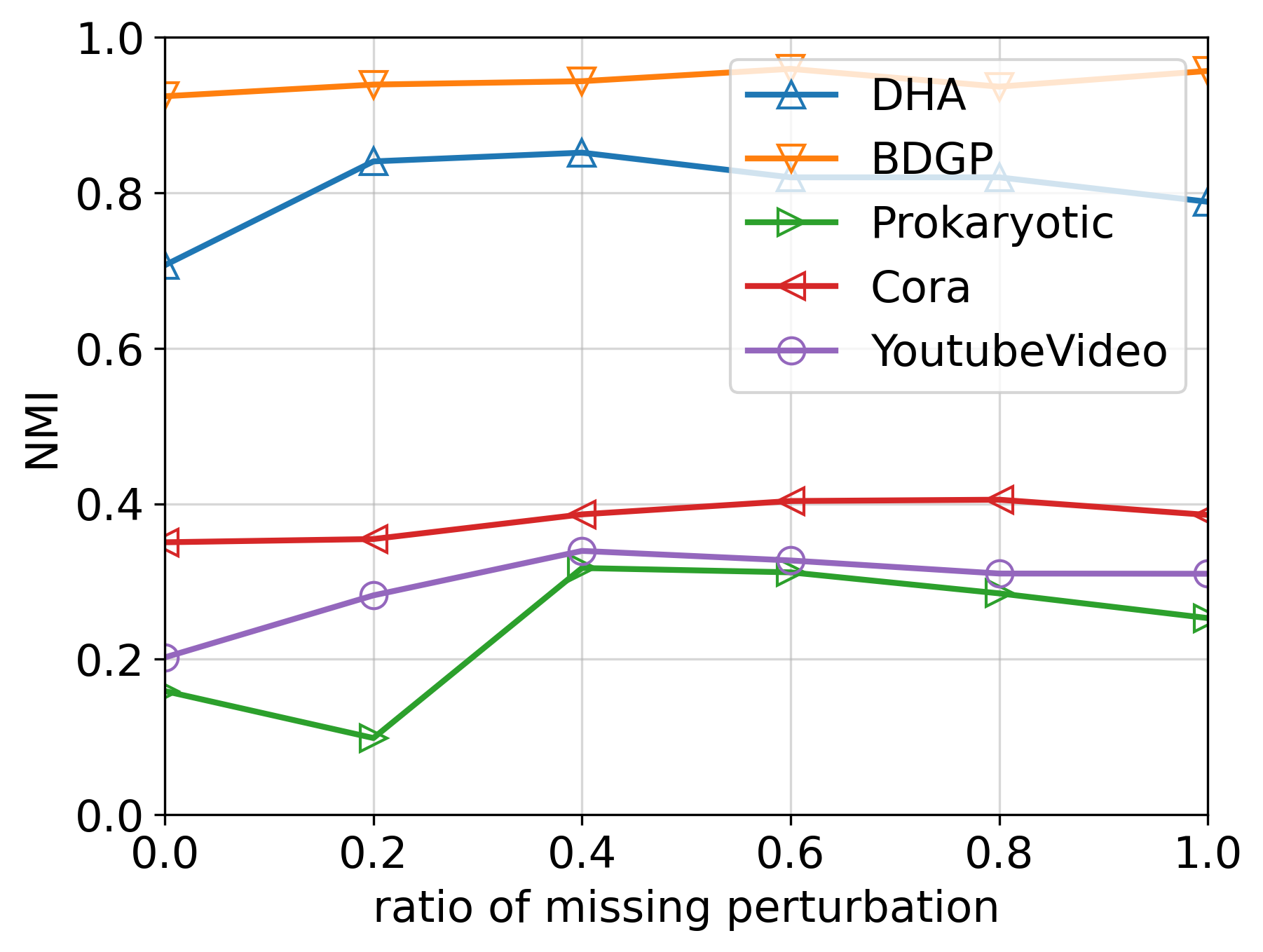}
\caption{$r$ $vs.$ NMI}
\end{subfigure}
\caption{Hyper-parameter analysis of the different ratios in our proposed simulated perturbation based multi-view contrastive learning on unsupervised multi-view clustering tasks, including noise perturbation (a-b) and unusable perturbation (c-d).}\label{PR}
\end{figure*}

\begin{figure*}[!ht]
\centering
\begin{subfigure}{0.19\linewidth}
\includegraphics[width=\linewidth]{Figure/DHA_loss.png}
\caption{DHA}
\end{subfigure}
\begin{subfigure}{0.19\linewidth}
\includegraphics[width=\linewidth]{Figure/BDGP_loss.png}
\caption{BDGP}
\end{subfigure}
\begin{subfigure}{0.19\linewidth}
\includegraphics[width=\linewidth]{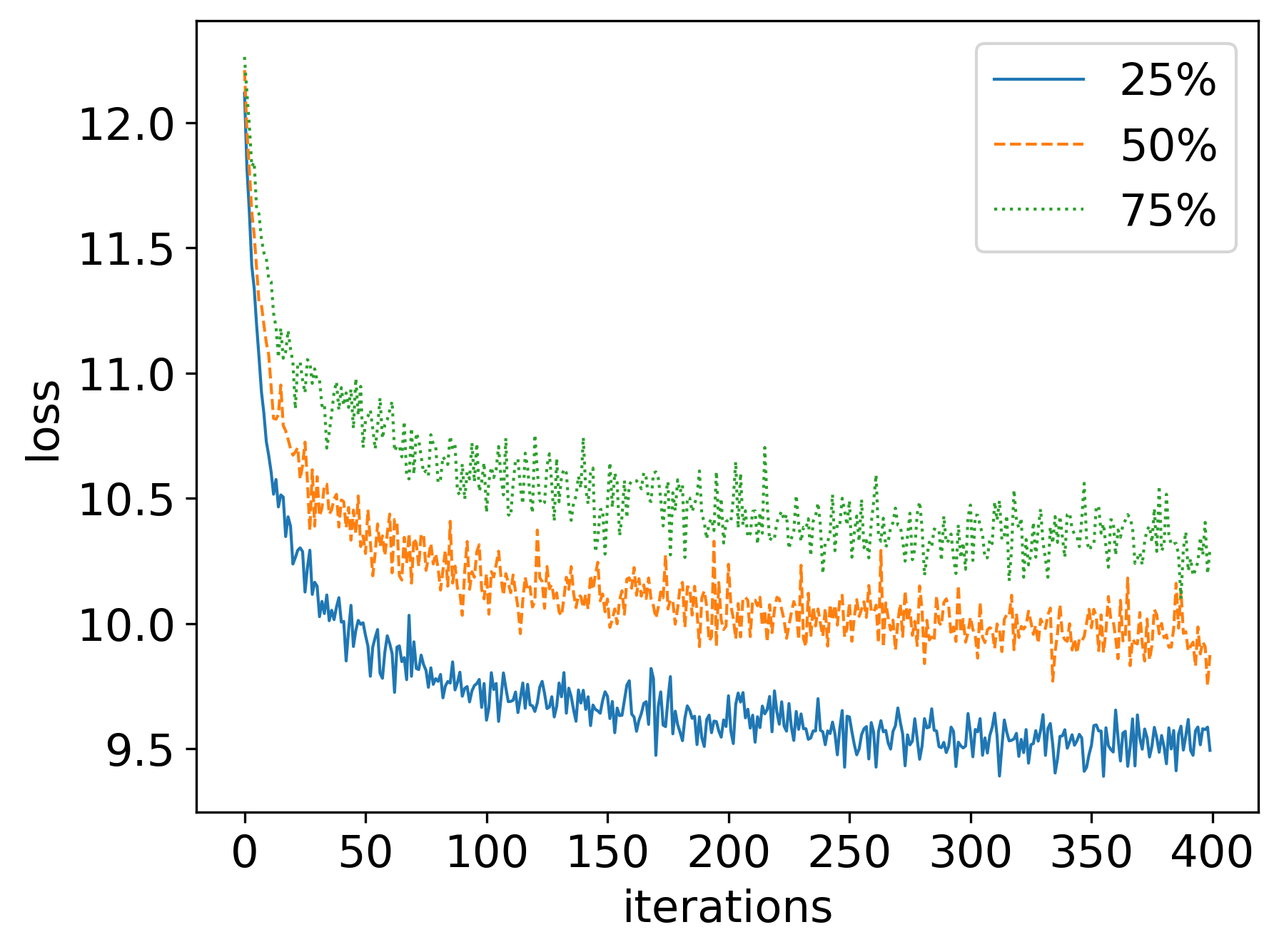}
\caption{Prokaryotic}
\end{subfigure}
\begin{subfigure}{0.19\linewidth}
\includegraphics[width=\linewidth]{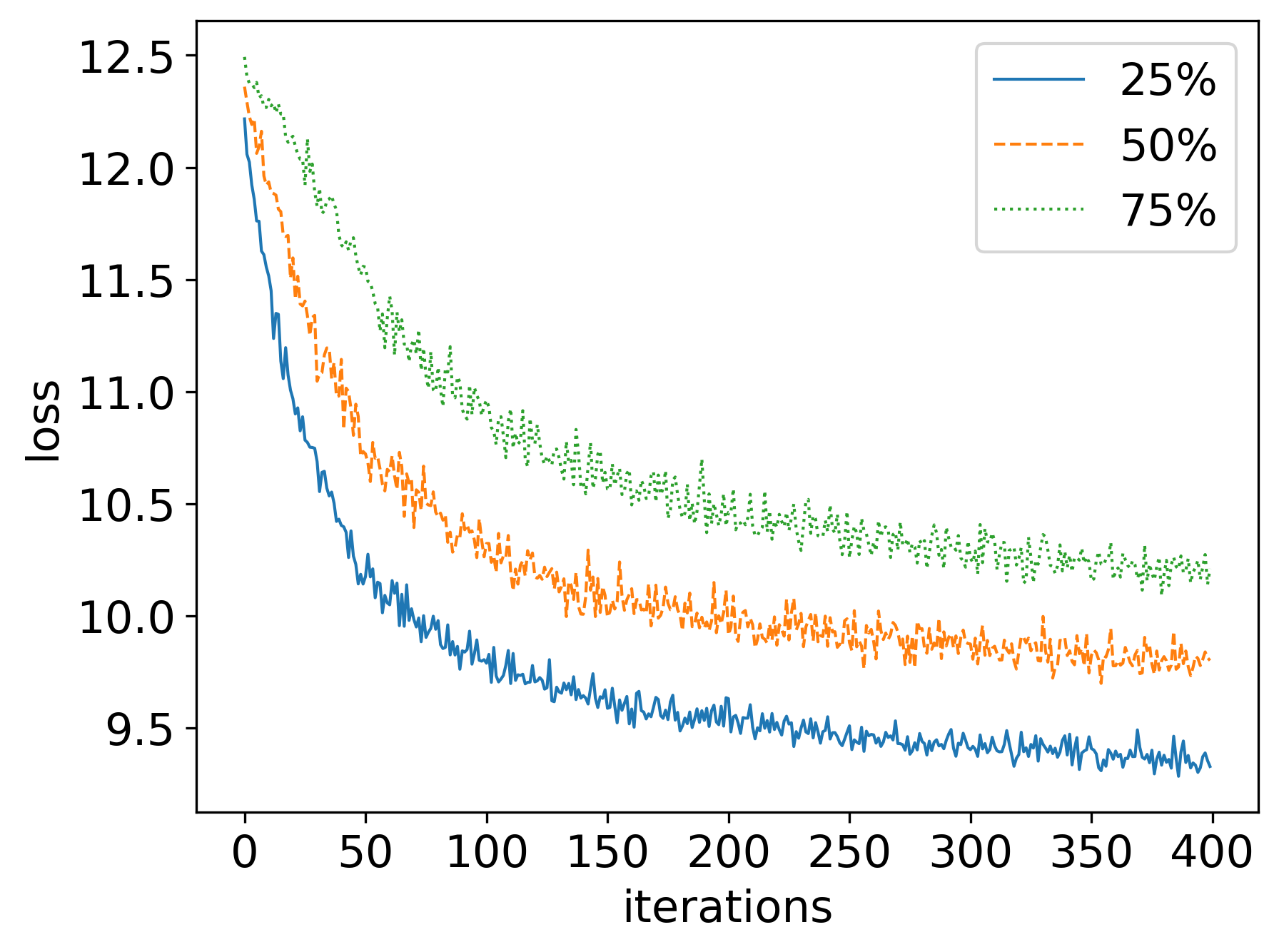}
\caption{Cora}
\end{subfigure}
\begin{subfigure}{0.19\linewidth}
\includegraphics[width=\linewidth]{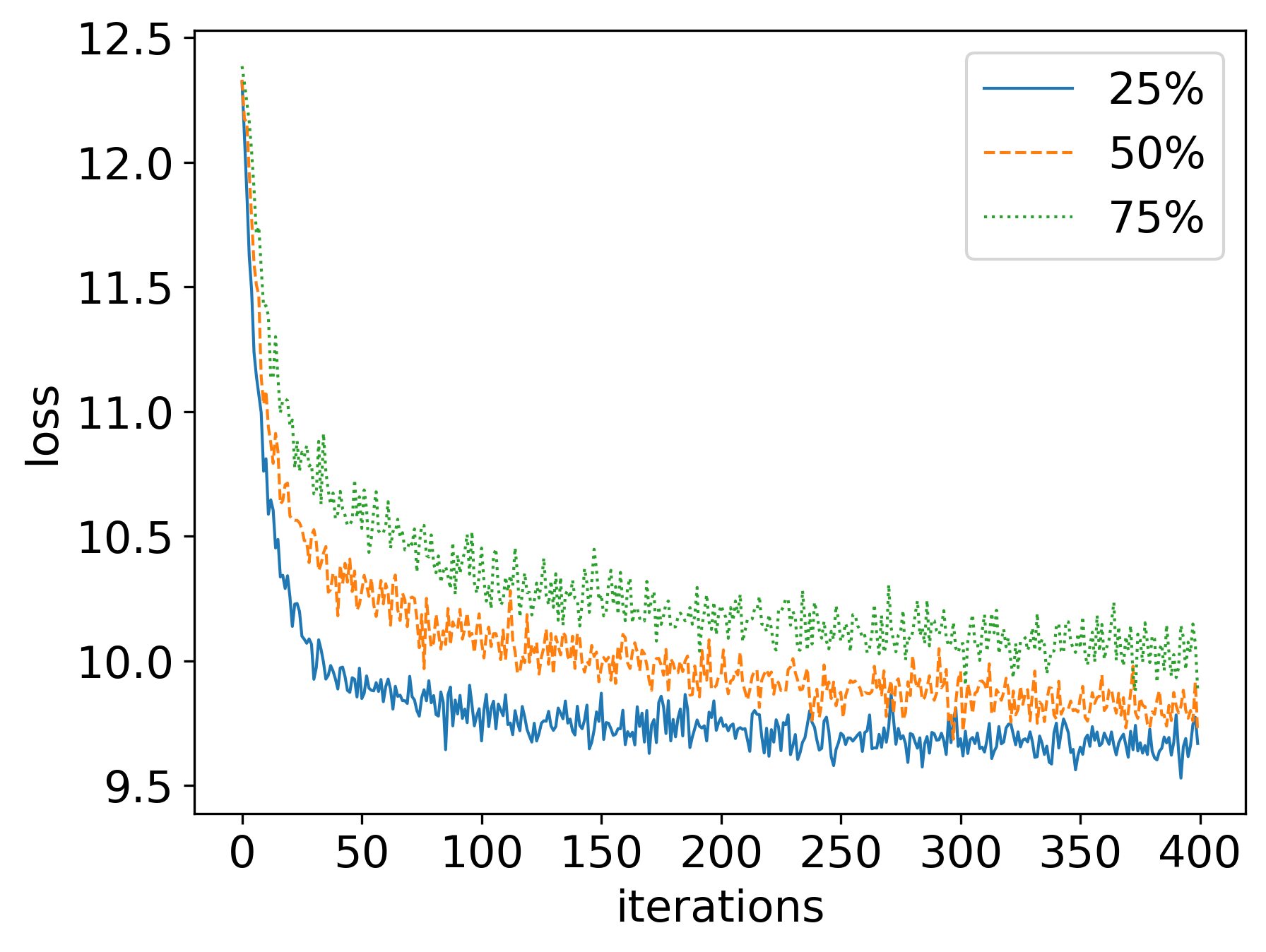}
\caption{YoutubeVideo}
\end{subfigure}
\caption{The training loss values during our proposed simulated perturbation based multi-view contrastive learning, indicating that RML has well-converged optimization objective even with different perturbation ratios ($25\%, 50\%, 75\%$).}\label{loss}
\end{figure*}

Figure~\ref{PR} and Figure~\ref{loss} provide additional visualization results on more datasets that are unable to be shown in the main paper due to space limitations.

\section{Potential Negative Societal Impacts}

In this paper, we propose a robust multi-view learning method, which works in the field of fundamental machine learning and computer vision algorithms. It will not produce new negative societal impacts beyond what we already know.

\end{document}